\documentclass[11pt]{article}

\usepackage[final]{acl}

\usepackage{times}
\usepackage{latexsym}

\usepackage[T1]{fontenc}
\usepackage[utf8]{inputenc}

\usepackage{microtype}

\usepackage{inconsolata}

\usepackage{graphicx}

\usepackage{kotex}
\usepackage{amsmath,amssymb,amsfonts}
\usepackage{subcaption}
\usepackage{multirow}
\usepackage{graphicx}
\usepackage{pifont}
\usepackage{listings} 
\usepackage{xcolor}
\usepackage{colortbl}
\usepackage{multirow}
\usepackage{booktabs}
\usepackage{xcolor}
\usepackage{newtxtext,newtxmath}
\usepackage{xcolor}
\usepackage{mdframed}
\mdfsetup{%
linecolor=white,
backgroundcolor=gray!20, 
}
\usepackage{authblk} 
\usepackage{hyperref} 
\usepackage{algorithm}
\usepackage{algpseudocode}

\title{A Linguistics-Aware LLM Watermarking via Syntactic Predictability}

\author{\textbf{Shinwoo Park}$^1$, \textbf{Hyejin Park}$^2$, \textbf{Hyeseon An}$^1$, \textbf{Yo-Sub Han}$^{1,\dagger}$ \\
  $^1$Yonsei University, Seoul, Republic of Korea \\
  \texttt{\{\href{mailto:pshkhh@yonsei.ac.kr}{pshkhh}, \href{mailto:hsan@yonsei.ac.kr}{hsan}, \href{mailto:emmous@yonsei.ac.kr}{emmous}\}@yonsei.ac.kr} \\
  $^2$Rensselaer Polytechnic Institute, Troy, NY, USA \\
  \href{mailto:parkh12@rpi.edu}{\texttt{parkh12@rpi.edu}} \\
}

\newcommand{\correspondingfootnote}{
    \let\oldthefootnote=\thefootnote
    \renewcommand{\thefootnote}{}
    \footnotetext{$\dagger$ Corresponding author.}
    \let\thefootnote=\oldthefootnote
}

\begin{document}

\maketitle

\correspondingfootnote 

\newcommand{\eg}{{\it e.g.}}
\newcommand{\ie}{{\it i.e.}}
\newcommand{\blue}[1]{\textcolor{blue}{\textbf{}#1}} 
\newcommand{\red}[1]{\textcolor{red}{\textbf{}(#1)}}
\newcommand{\green}[1]{\textcolor{green}{\textbf{}#1}}

\begin{abstract}

As large language models~(LLMs) continue to advance rapidly, 
reliable governance tools have become critical. 
Publicly verifiable watermarking is particularly essential 
for fostering a trustworthy AI ecosystem.
A central challenge persists: balancing text quality against detection robustness. 
Recent studies have sought to navigate this trade-off by leveraging signals 
from model output distributions~(e.g., token-level entropy); however, 
their reliance on these model-specific signals presents a significant barrier 
to public verification, 
as the detection process requires access to the logits of the underlying model.
We introduce STELA, a novel framework that aligns watermark strength with the linguistic degrees 
of freedom inherent in language. 
STELA dynamically modulates the signal using part-of-speech~(POS) n-gram–modeled 
linguistic indeterminacy, 
weakening it in grammatically constrained contexts 
to preserve quality and 
strengthening it in contexts with greater linguistic flexibility to enhance detectability.
Our detector operates without access to 
any model logits, 
thus facilitating publicly verifiable detection. 
Through extensive experiments on typologically diverse languages—analytic English, 
isolating Chinese, and agglutinative Korean—we 
show that STELA surpasses prior methods in detection robustness.
Our code is available at \url{https://github.com/Shinwoo-Park/stela_watermark}.

\end{abstract}

\section{Introduction}
\label{sec:introduction}

The rapid advancement of LLMs has demonstrated remarkable capabilities across 
a wide range of applications, 
spanning from creative writing~\citep{gomez-rodriguez-williams-2023-confederacy} and 
generating articles at the level of Wikipedia entries from scratch~\citep{shao-etal-2024-assisting}
to operating in specialized domains such as law~\citep{hu-etal-2025-fine} 
and medical question answering~\citep{labrak-etal-2024-biomistral}.
However, this progress raises concerns over their potential 
for misuse~\citep{mitchell2023detectgpt,yang-etal-2024-survey,park-etal-2025-katfishnet}. 
The propensity of LLMs to hallucinate facts~\citep{jiang-etal-2024-large} 
and their capacity to generate convincing disinformation at scale~\citep{vykopal-etal-2024-disinformation} 
pose significant societal risks. 
In response, the EU AI Act 
requires providers to make the outputs of 
general-purpose AI models clearly identifiable~\citep{euaiact2024}. 

LLM watermarking~\citep{liu2024survey,lalai-etal-2025-intentions,kim2025marking,an-etal-2026-ditto,park2026watermod} 
has emerged as a promising approach to address these challenges. 
A representative scheme, KGW, proposed by \citet{kirchenbauer2023watermark}, 
partitions the model vocabulary into green and red lists at each generation step, 
softly promoting the selection of green-listed tokens to embed a statistical signal. 
A key strength of this approach is that a third party can detect the watermark 
without requiring access to the model parameters, thereby enabling public verification.
However, the authors noted a trade-off between text quality 
and detection robustness~\citep{kirchenbauer2023watermark,giboulot2024watermax}. 
In contexts with low token-level entropy, where one token is far more probable than others, 
biasing the logits for green tokens may be insufficient to alter the output. 
Forcing the selection of a green token in such cases risks degrading text quality 
by generating an unnatural word. 

This limitation motivated a new wave of adaptive methods that modulate watermark strength.  
SWEET~\citep{lee-etal-2024-wrote} selectively applies watermarking only at 
generation steps where the 
token entropy exceeds a predefined threshold, 
and its detector confines analysis to these high-entropy positions. 
EWD~\citep{lu-etal-2024-entropy} assigns each token 
a weight 
proportional to its entropy in the final statistic, 
reducing 
the influence of less reliable, 
low-entropy tokens without a hard threshold.
While effective, these methods introduce a significant obstacle to public verifiability. 
Since both SWEET and EWD require token entropy for detection, 
they necessitate access to the logits of the LLM. 

We propose a novel framework that aligns the watermark signal 
with the linguistic degrees of freedom
inherent in language. 
Our method, STELA~(\textbf{S}tructurally-\textbf{T}ethered \textbf{E}ntropy-based \textbf{L}inguistic w\textbf{A}termarking), 
is the first to modulate watermark strength for both 
insertion and detection based on linguistic indeterminacy, 
a concept we model using the conditional entropy of POS n-grams. 
This approach distinguishes between two sources of low token entropy: 
semantic fixedness~(e.g., a proper noun) 
and grammatical necessity~(e.g., a mandatory particle).
It weakens the signal in grammatically constrained contexts to preserve quality 
and strengthens the signal where linguistic choices are abundant to enhance detectability. 
A primary strength of this design is that the detection process does not require access 
to any model logits; like the original KGW framework, 
it enables public verification using only a lightweight POS analyzer.
% New! Camera-ready 
Critically, whereas entropy-based methods treat the transient uncertainty of the model as their reference signal, 
STELA treats the intrinsic syntactic structure of the language as its reference. 
This shift from model-specific to language-universal grounding enables robust public auditability, 
a property that token-entropy methods inherently lack.

We validate the generalizability of STELA across three distinct language types: 
analytic English, where grammatical relations are primarily conveyed by strict word 
order and function words; 
isolating Chinese, an extreme analytic case where word order is paramount and inflection 
is virtually absent; 
and agglutinative Korean, where rich morphological suffixes~(e.g., particles and endings) 
are attached to lexical stems to encode grammar, allowing for flexible word order. 
Our experiments demonstrate that STELA achieves 
superior watermark detection performance compared to prior methods.

\begin{table*}[hbt!]
\centering
\resizebox{\textwidth}{!}{
\begin{tabular}{l l l l l}
\toprule
\textbf{Method} & \textbf{Insertion Strategy} & \textbf{Detection Strategy} & \textbf{Detection framework} & \textbf{Adaptation Signal} \\
\midrule
KGW~\citep{kirchenbauer2023watermark} & Static Strength & Uniform Count & Model-Free & N/A \\
SWEET~\citep{lee-etal-2024-wrote} & Selective~(Threshold-based) & Selective~(Threshold-based) & Model-Dependent & Token Entropy \\
EWD~\citep{lu-etal-2024-entropy} & Static Strength & Adaptive~(Weighted) & Model-Dependent & Token Entropy \\
MorphMark~\citep{wang-etal-2025-morphmark} & Adaptive~(Weighted) & Uniform Count & Model-Free & Green Token Probability \\
\midrule
\textbf{STELA~(Ours)} & \textbf{Adaptive~(Weighted)} & \textbf{Adaptive~(Weighted)} & \textbf{Model-Free} & \textbf{Linguistic Indeterminacy} \\
\bottomrule
\end{tabular}
}
\caption{
A taxonomy of representative watermarking methods. 
STELA integrates a fully adaptive strategy for 
both insertion and detection 
with the model-free detection framework. 
Its key distinction from prior adaptive methods is 
the use of a model-independent, 
linguistic indeterminacy signal, rather than model-specific logits.
}\label{tab:watermark_taxonomy}
\end{table*}

\section{Related Work}

Table~\ref{tab:watermark_taxonomy} provides a taxonomy of representative LLM watermarking methods.
Much of the current research builds on the foundational work of
KGW~\citep{kirchenbauer2023watermark}.
This pioneering method established a practical framework for logit-based watermarking. 
It partitions the vocabulary into green and red lists based on a hash of the preceding token 
and applies a static bias to the logits of green-listed tokens during generation. 
A key advantage is its model-free detection, 
allowing third parties to verify the watermark 
using only a public hash function and a statistical z-test.
MorphMark~\citep{wang-etal-2025-morphmark} builds upon this foundation by 
introducing an adaptive insertion strategy while 
preserving the model-free detection of KGW. 
It dynamically adjusts watermark strength based on the cumulative probability of the 
green-listed tokens at each generation step. 
While its detection remains publicly verifiable, 
the insertion logic is still guided by the transient output probabilities of the model. 

Another line of work tackles the challenge of watermarking in low-entropy contexts, 
where a static bias is often insufficient to alter a sharply peaked token distribution. 
This can lead to failed watermark insertions and subsequent detection failures. 
Recent methods address this challenge by incorporating token-level entropy 
calculated from model logits.
SWEET~\citep{lee-etal-2024-wrote}, 
for instance, selectively applies the watermark only when token entropy exceeds 
a predefined threshold, confining both insertion and detection to these positions. 
EWD~\citep{lu-etal-2024-entropy} 
takes a different approach by proposing an entropy-aware detection method that 
assigns a continuous weight to each token in the final statistic, proportional to its entropy, 
thereby reducing the influence of less reliable low-entropy tokens. 
Although these methods provide refined adaptations, they simultaneously impose  
an obstacle to public verifiability. 
Specifically, because both SWEET and EWD rely on token-level entropy for 
their detection mechanisms, 
they are model-dependent and therefore require direct access to the 
LLM logits.

Further discussion of watermarking methods is provided
in Appendix~\ref{appendix:related_work}.

\section{Background}

\paragraph{Next-Token Generation.}
At each generation step $t$, an LLM takes a sequence of preceding tokens 
$x_{<t} = (x_1, \dots, x_{t-1})$ as input. 
The model then computes a logit vector $l_t \in \mathbb{R}^{|\mathcal{V}|}$ 
over the entire vocabulary $\mathcal{V}$. 
This vector represents the unnormalized log-probabilities for 
the next token over the entire vocabulary.
The logits are converted into a probability distribution $p_t$ via the softmax function:
\begin{equation}
    p_{t,i} = \frac{\exp(l_{t,i})}{\sum_{j=1}^{|\mathcal{V}|} \exp(l_{t,j})},
    \label{eq:softmax}
\end{equation}
where $p_{t,i}$ is the probability of the $i$-th token. 
The next token $x_t$ is then sampled from this distribution.

\paragraph{Watermark Insertion.}
The KGW watermarking scheme intervenes in the generation process immediately before the sampling step.
First, at each step $t$, a public pseudorandom function, typically a hash function, 
is applied to the preceding token $x_{t-1}$ to generate a seed. 
This seed is used to deterministically partition the entire vocabulary $\mathcal{V}$ 
into a green list $\mathcal{V}_G$ and a red list $\mathcal{V}_R$. 
The size of the green list is determined by a predefined ratio $\gamma$, 
such that $|\mathcal{V}_G| = \gamma |\mathcal{V}|$.
Next, the method modifies the logit vector $l_t$ by adding a fixed positive bias $\delta$ 
to the logits of all tokens belonging to the green list. 
The modified logit for the $i$-th token, $l'_{t,i}$, is given by:
\begin{equation}
    l'_{t,i} = 
    \begin{cases} 
        l_{t,i} + \delta & \text{if } i \in \mathcal{V}_G \\
        l_{t,i} & \text{if } i \in \mathcal{V}_R 
    \end{cases}
    \label{eq:logit_modification}
\end{equation}
A new, watermarked probability distribution $p'_t$ is then computed by applying the 
softmax function to the modified logits $l'_t$. 
The final token $x_t$ is sampled from this biased distribution $p'_t$,
which encourages the selection of green-listed tokens.

\paragraph{Watermark Detection.}
Given a text sequence $X = (x_1, \dots, x_T)$, 
the detector first replicates the green list generation process. 
For each token $x_t$~(from $t=2$ to $T$), 
it applies the same public hash function 
to the preceding token $x_{t-1}$ to reconstruct the green list 
$\mathcal{V}_G$ for that step.
The null hypothesis, $H_0$, is that the text was generated without knowledge of the green list rule.
Under $H_0$, the probability of any token being in its corresponding green list is 
simply the green list ratio, $\gamma$. 
For a text of total length $T$, let $|s|_G$ denote the number of tokens that fall into their respective green lists.
A z-score is then calculated using the one-proportion z-test formula:
\begin{equation}
    z = \frac{|s|_G - \gamma T}{\sqrt{T\gamma(1-\gamma)}}.
    \label{eq:z_score}
\end{equation}
This score measures how many standard deviations the observed count of green tokens 
is from the expected count under the null hypothesis. 
If the calculated z-score exceeds a predefined threshold~(e.g., $z > 4.0$), the text is identified as watermarked.

\begin{figure*}[hbt!]
\includegraphics[width=\textwidth]{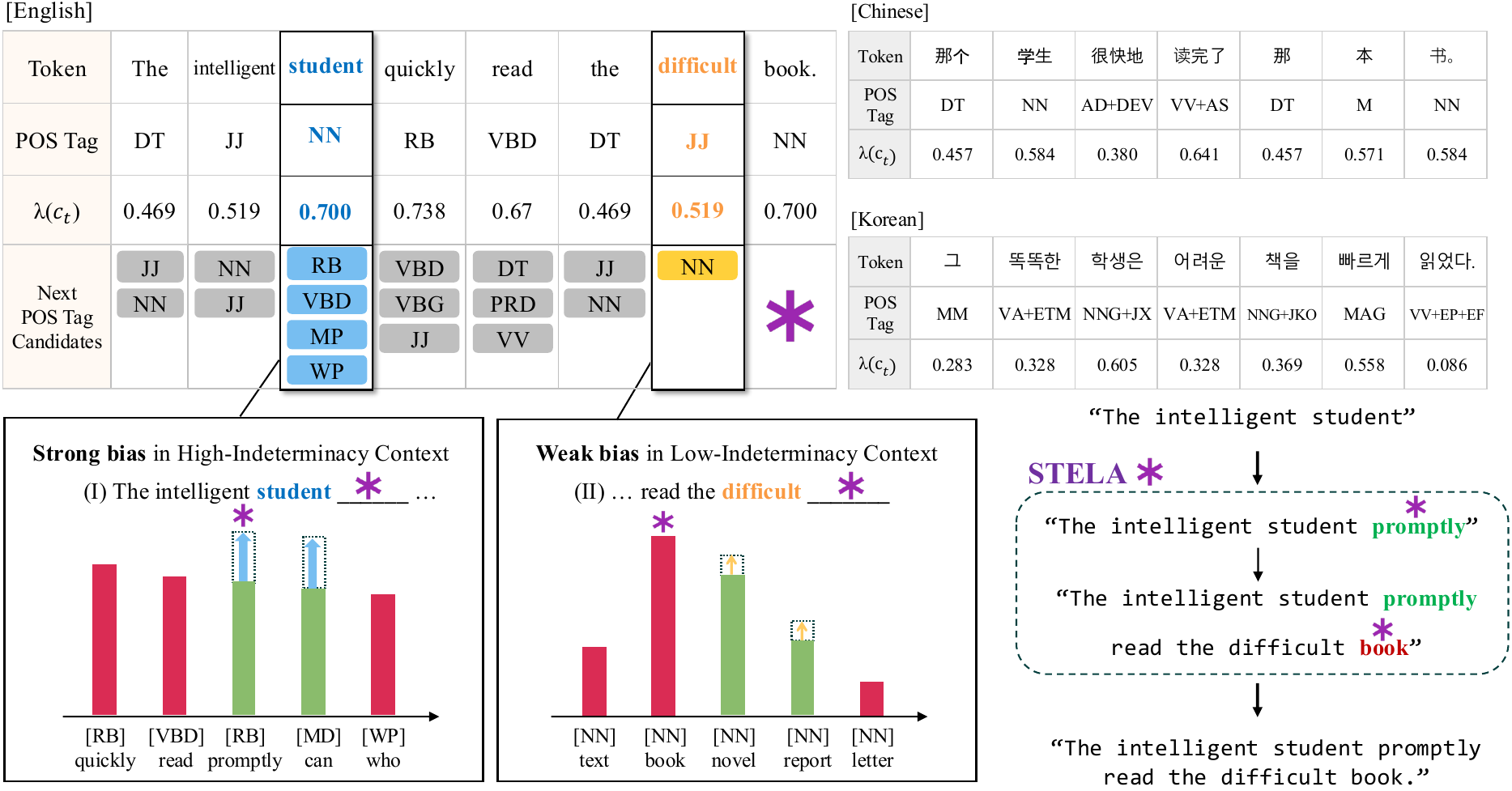}
    \caption{Adaptive watermark insertion by linguistic indeterminacy. 
    }\label{fig:stela_figure}
\end{figure*}

\section{Method}
\label{sec:method}

We first define a model-independent measure of linguistic indeterminacy. 
Then, we detail how this measure informs novel adaptive strategies 
for both watermark insertion and detection.
The pseudocode for STELA is provided in Appendix~\ref{appendix:algorithms}.

\subsection{Defining Linguistic Indeterminacy}
\label{sec:linguistic_indeterminacy}

Whereas prior adaptive methods rely on model-dependent 
\textit{probabilistic uncertainty}~(i.e., token entropy), 
STELA introduces a new framework based on model-independent 
\textit{linguistic indeterminacy}. 
We formalize this concept by quantifying 
the predictability of syntactic structures using part-of-speech n-grams. 
Specifically, we define a measure, $\lambda(c_t)$, 
that captures the degree of grammatical freedom at a given generation step~$t$.

Let $\mathcal{S} = (\pi_1, \dots, \pi_T)$ 
be the sequence of POS tags corresponding to a text sequence. 
Here, $k$ is a hyperparameter representing the n-gram size of the POS context.
At step $t$, the linguistic context is defined as a sequence
of the preceding $k-1$ POS tags, 
denoted by $c_t = (\pi_{t-k+1}, \dots, \pi_{t-1})$.
We estimate the conditional probability distribution $P(\pi_t | c_t)$ 
of the next POS tag $\pi_t$ given the context $c_t$ from 
a human-written corpus. 
The conditional Shannon entropy is defined as:
\begin{equation}
    H(P(\pi_t | c_t)) = - \sum_{\pi' \in \mathcal{V}_{\text{POS}}} P(\pi' | c_t) \log P(\pi' | c_t),
    \label{eq:shannon_entropy}
\end{equation}
where $\mathcal{V}_{\text{POS}}$ represents the set of all possible POS tags. 
The linguistic indeterminacy $\lambda(c_t)$ is then defined as the 
normalized entropy of this distribution:
\begin{equation}
    \lambda(c_t) = \frac{H(P(\pi_t | c_t))}{\log K_{c_t}},
    \label{eq:lambda}
\end{equation}
where $K_{c_t}$ is the number of unique tag types observed following 
the context $c_t$ in the corpus.
A value of $\lambda(c_t) \to 1$ signifies 
high indeterminacy~(many grammatical options are plausible), 
whereas $\lambda(c_t) \to 0$ indicates 
high predictability~(the syntactic structure is strongly constrained). 
This pre-computed, model-agnostic signal serves as the foundation 
for our approach.
During both watermark insertion and detection, 
STELA references this signal via a lookup table to retrieve the 
$\lambda(c_t)$ value corresponding to the current context $c_t$.
 
\subsection{Watermark Insertion: Adaptive Strength}
\label{sec:stela_insertion}

STELA refines 
the static insertion strategy of KGW
by dynamically modulating the watermark strength based on 
linguistic indeterminacy. 
Instead of applying a fixed bias $\delta$, 
STELA computes an adaptive bias $\delta'_t$ for each generation step $t$:
\begin{equation}
    \delta'_t = \delta \cdot \lambda(c_t),
    \label{eq:adaptive_delta}
\end{equation}
where $c_t$ is the POS context of the preceding tokens. 
The logit modification for the $i$-th token in the vocabulary is then:
\begin{equation}
    l'_{t,i} = 
    \begin{cases} 
        l_{t,i} + \delta'_t & \text{if } i \in \mathcal{V}_G \\
        l_{t,i} & \text{if } i \in \mathcal{V}_R 
    \end{cases}
    \label{eq:stela_logit_modification}
\end{equation}
This mechanism applies a strong watermark bias in linguistically 
flexible contexts~(where $\lambda(c_t)$ is high) 
and a weak or negligible bias in grammatically 
constrained contexts~(where $\lambda(c_t)$ is low). 

Figure~\ref{fig:stela_figure} illustrates the adaptive watermark insertion of STELA.

\subsection{Watermark Detection: Adaptive Scoring}
\label{sec:stela_detection}

Corresponding to the adaptive insertion, 
the detection process of STELA replaces the uniform counting of KGW 
with an adaptive scoring mechanism. 
The contribution of each token to the detection score is weighted by 
its linguistic indeterminacy. 
For each token $x_t$ in a sequence of length $T$, we define its weight 
$w_t$ as:
\begin{equation}
    w_t = \lambda(c_t),
    \label{eq:weight}
\end{equation}
where $c_t$ is the POS context reconstructed from the preceding tokens.
We then formulate a weighted z-score, $z'$, 
to test the null hypothesis $H_0$. 
The observed statistic is the sum of weights of the green-listed tokens, 
$W_G = \sum_{t=1}^{T} w_t \cdot \mathbb{I}(x_t \in \mathcal{V}_{G,t})$, 
where $\mathbb{I}(\cdot)$ is the indicator function. 
Under $H_0$, the expected value of 
$W_G$ is $\mathbb{E}[W_G] = \gamma \sum_{t=1}^{T} w_t$, 
and its variance is $\text{Var}(W_G) = \gamma(1-\gamma) \sum_{t=1}^{T} w_t^2$.
The final $z'$ score is thus calculated as:
\begin{equation}
    z' = \frac{\sum_{t=1}^{T} w_t \cdot \mathbb{I}(x_t \in \mathcal{V}_{G,t}) - \gamma \sum_{t=1}^{T} w_t}{\sqrt{\gamma(1-\gamma) \sum_{t=1}^{T} w_t^2}}.
    \label{eq:weighted_z_score}
\end{equation}
This adaptive scoring mechanism increases the weight of tokens 
in high-indeterminacy contexts, where the watermark is intentionally 
embedded with greater strength, 
and reduces the impact of tokens in more predictable contexts.
Consequently, the detection statistic exhibits higher sensitivity 
and robustness compared to the uniform counting baseline.
In line with KGW, the entire procedure remains model-free.

\section{Experimental Setup}
\label{sec:experimental_setup}

\subsection{Measuring Linguistic Indeterminacy}
\label{sec:measure_linguistic_indeterminacy}

The statistical reliability of the linguistic indeterminacy signal, 
$\lambda(c_t)$, depends on the breadth and quality of the 
human-written reference corpora. 
We estimate POS n-gram statistics from two corpora per language.
For English, we use the English subset of the Wikipedia~\citep{wikimedia2023wikipedia} dataset 
and OpenWebText2~\citep{geralt2025openwebtext2} dataset. 
For Korean, 
we use the Korean subset of the C4 dataset~\citep{allenai2019c4} and 
KOREAN-WEBTEXT~\citep{haerae2024koreanwebtext} dataset. 
For Chinese, we use the Chinese subset of the Wikipedia~\citep{wikimedia2023wikipedia} dataset 
and the Chinese subset of the C4 dataset~\citep{allenai2019c4}. 
From each corpus, we randomly sample 10,000 documents. 
We derive the final $\lambda$ table for each language by 
averaging the conditional probabilities computed from the two 
respective corpora, which mitigates any potential bias from a single source. 
We employ standard POS taggers for each language: 
spaCy~(Penn Treebank tagset)~\citep{spacy2020} for English, 
HanLP~(Penn Chinese Treebank tagset) for Chinese~\citep{he-choi-2021-stem},
and Kiwi~(Sejong tagset) for Korean~\citep{lee2024kiwi}.
A description of these taggers and their tagsets 
is provided in Appendix~\ref{appendix:language_pos}.

The choice of the POS context size, $k$, 
requires balancing the need for sufficient contextual information against 
the risk of data sparsity. 
Our selection of different $k$ values for each language is a deliberate 
choice informed by their typological characteristics. 
For analytic English, which employs a relatively strict word order, 
a local context of $k=2$ proves effective. 
For agglutinative Korean and isolating Chinese, where phrasal structures 
and dependencies can be more complex, we use a larger context of $k=4$ 
to better capture the syntactic state. 
In Section~\ref{sec:pos_context_length}, we analyze the performance variation 
with respect to the value of $k$.

\subsection{Main Evaluation Settings} 
\label{sec:main_evaluation}

\paragraph{Evaluation Datasets and Tasks.}
We select a distinct task for each language, using 500 randomly sampled instances 
from each dataset. 
For English, we use the WritingPrompts dataset~\citep{fan-etal-2018-hierarchical} 
to test creative story continuation, 
using the story prompts provided within the dataset as the initial context. 
For Chinese, we use the THUCNews dataset~\citep{thuc2023news} to test 
fact-based news article continuation. 
For Korean, we use the Korean subset of the Oscar 
dataset~\citep{ortiz-suarez-etal-2020-monolingual} to evaluate general-purpose web 
text continuation. 
For both the Chinese and Korean tasks, we use the first three sentences as prompts.
This diversity in tasks supports the generalizability of our findings beyond 
a single style of text generation.

\paragraph{Models.} 
Three principles guide the selection of LLMs for this study: 
regional representation, 
comparable scale, and architectural diversity. 
We choose three open-source LLMs of a similar scale~(0.5B–1B parameters) 
to facilitate a fair comparison while reducing potential 
confounding effects arising from model size.
This selection of models represents distinct AI ecosystems: 
Llama-3.2-1B~\citep{meta2024llama3.2} from the US, 
advancing open-source foundation models; Qwen3-0.6B~\citep{yang2025qwen3} 
from China, contributing to multilingual LLM research; 
and HyperCLOVAX-SEED-Text-0.5B~\citep{naver2025hyperclovaXseed} 
from South Korea, focusing on regionally optimized models.
This diversity also serves to validate the generalizability 
of STELA across different model architectures.

\subsection{Evaluation Metrics}
For detection performance, we employ two metrics: 
the true positive rate at a 5\% false positive rate~(TPR@5\%FPR), which reflects a practical scenario where minimizing false accusations 
is critical, and the best F1 score, 
representing the optimal balance between precision and recall achieved by selecting the best detection threshold. 
We measure text quality using perplexity~(PPL). 
A lower PPL indicates higher fluency,
which we calculate for each generated text 
using Llama-3.2-3B as an external reference model.

\begin{table*}[hbt!]
\centering\small
\begin{tabular}{llcccccc}
\toprule
\multirow{2}{*}{LLM} & \multirow{2}{*}{Method} & \multicolumn{2}{c}{English} & \multicolumn{2}{c}{Chinese} & \multicolumn{2}{c}{Korean} \\ 
\cmidrule(lr){3-4} \cmidrule(lr){5-6} \cmidrule(lr){7-8}
& & TPR@FPR5\% & Best F1 & TPR@FPR5\% & Best F1 & TPR@FPR5\% & Best F1 \\ 
\midrule

\multirow{5}{*}{Llama-3.2}
& KGW & \textbf{0.950} & \textbf{0.963} & 0.962 & 0.963 & 0.906 & 0.932 \\
& SWEET & 0.850 & 0.906 & 0.872 & 0.910 & 0.862 & 0.912 \\
& EWD & 0.870 & 0.916 & 0.850 & 0.902 & 0.896 & 0.928 \\
& MorphMark & 0.926 & 0.943 & 0.936 & 0.945 & 0.826 & 0.893 \\
& STELA & 0.938 & 0.953 & \textbf{0.976} & \textbf{0.972} & \textbf{0.950} & \textbf{0.954} \\ 
\cmidrule(lr){1-8} 

\multirow{5}{*}{Qwen-3}
& KGW & 0.958 & 0.957 & 0.992 & 0.990 & 0.902 & 0.934 \\
& SWEET & 0.968 & 0.967 & 0.990 & 0.990 & 0.914 & 0.931 \\
& EWD & \textbf{0.986} & \textbf{0.978} & 0.992 & 0.990 & 0.914 & 0.934 \\
& MorphMark & 0.870 & 0.942 & 0.976 & 0.978 & 0.778 & 0.892 \\
& STELA & 0.978 & 0.966 & \textbf{0.996} & \textbf{0.994} & \textbf{0.950} & \textbf{0.950} \\ 
\cmidrule(lr){1-8} 

\multirow{5}{*}{HyperCLOVA}
& KGW & 0.978 & 0.970 & 0.928 & 0.939 & 0.948 & 0.955 \\
& SWEET & 0.962 & 0.964 & 0.850 & 0.904 & 0.916 & 0.935 \\
& EWD & 0.962 & 0.964 & 0.858 & 0.908 & 0.952 & 0.957 \\
& MorphMark & 0.910 & 0.934 & 0.756 & 0.870 & 0.766 & 0.875 \\
& STELA & \textbf{0.988} & \textbf{0.975} & \textbf{0.932} & \textbf{0.942} & \textbf{0.960} & \textbf{0.960} \\ 

\bottomrule
\end{tabular}
\caption{
A comparison of detection performance among various watermarking methods. 
The best-performing results are highlighted in bold. 
For each of the three LLMs, 
STELA achieves the highest average detection performance across 
the three languages.
}
\label{tab:detection_results}
\end{table*}

\begin{figure*}[hbt!]
    \centering

    \begin{subfigure}{0.32\textwidth}
        \includegraphics[width=\linewidth]{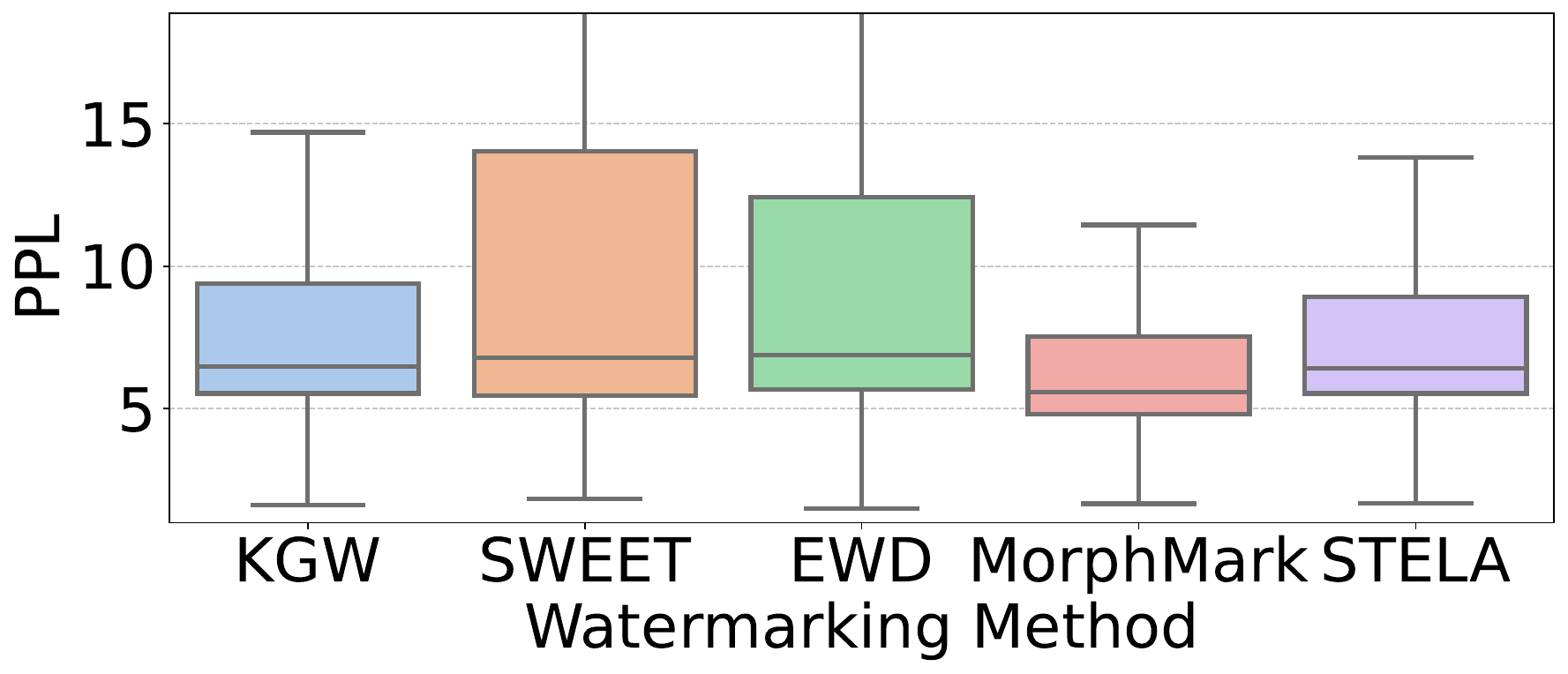}
    \end{subfigure}\hfill
    \begin{subfigure}{0.32\textwidth}
        \includegraphics[width=\linewidth]{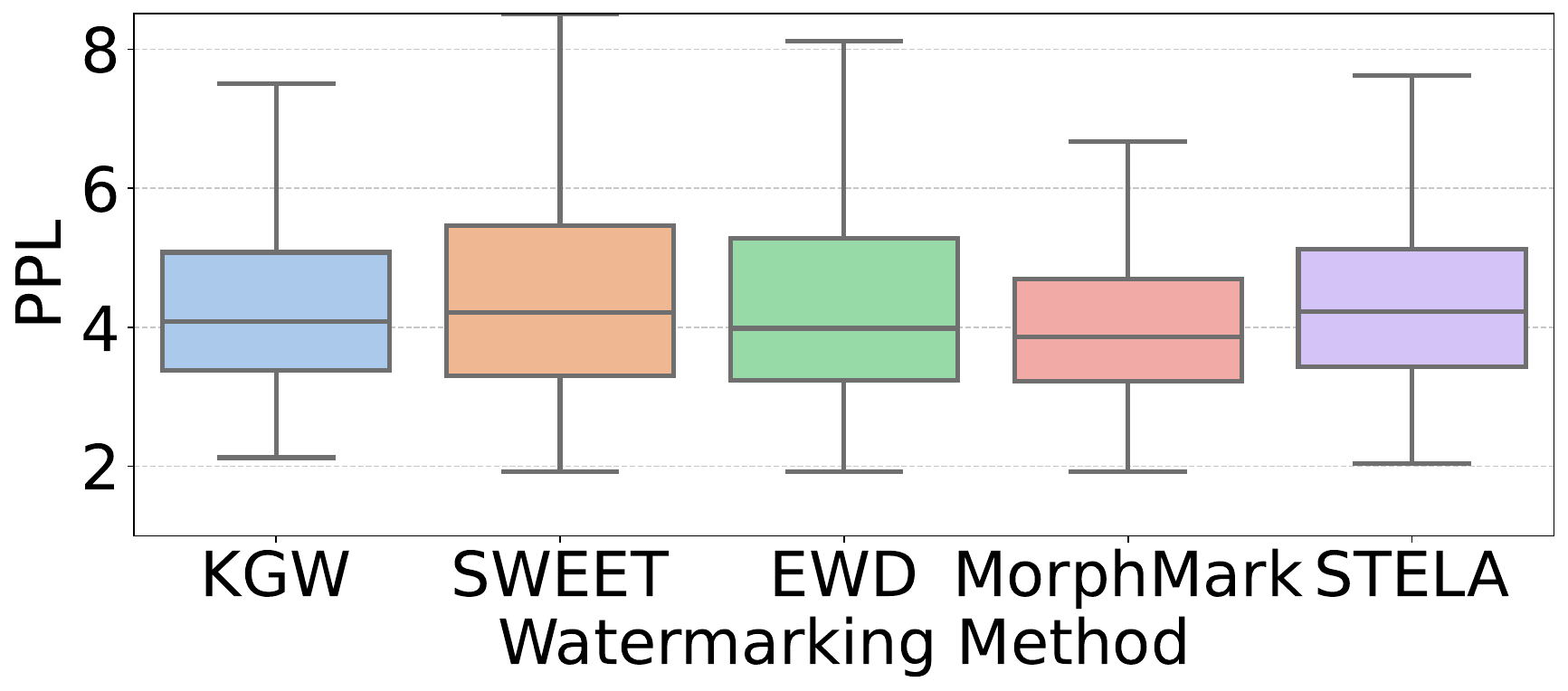}
    \end{subfigure}\hfill
    \begin{subfigure}{0.32\textwidth}
        \includegraphics[width=\linewidth]{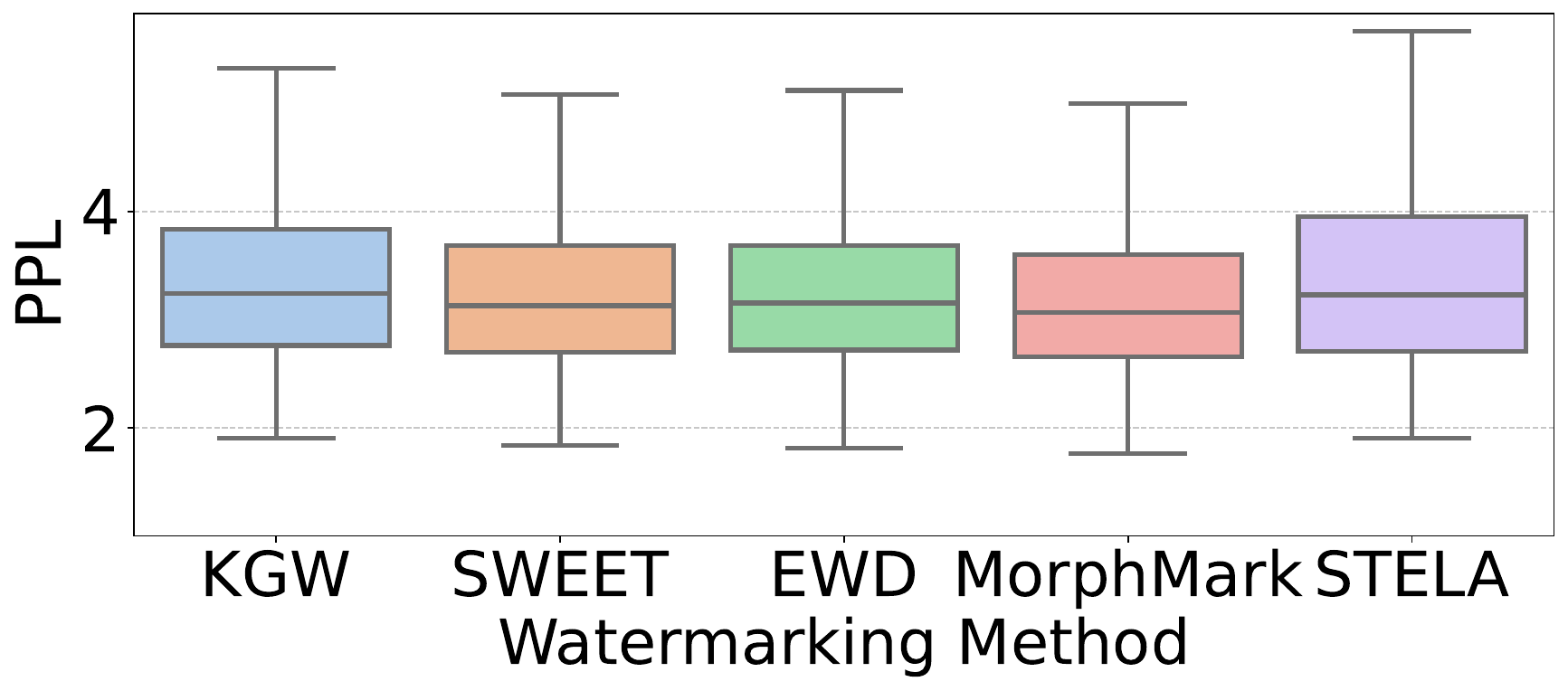}
    \end{subfigure}
    
    \begin{subfigure}{0.32\textwidth}
        \includegraphics[width=\linewidth]{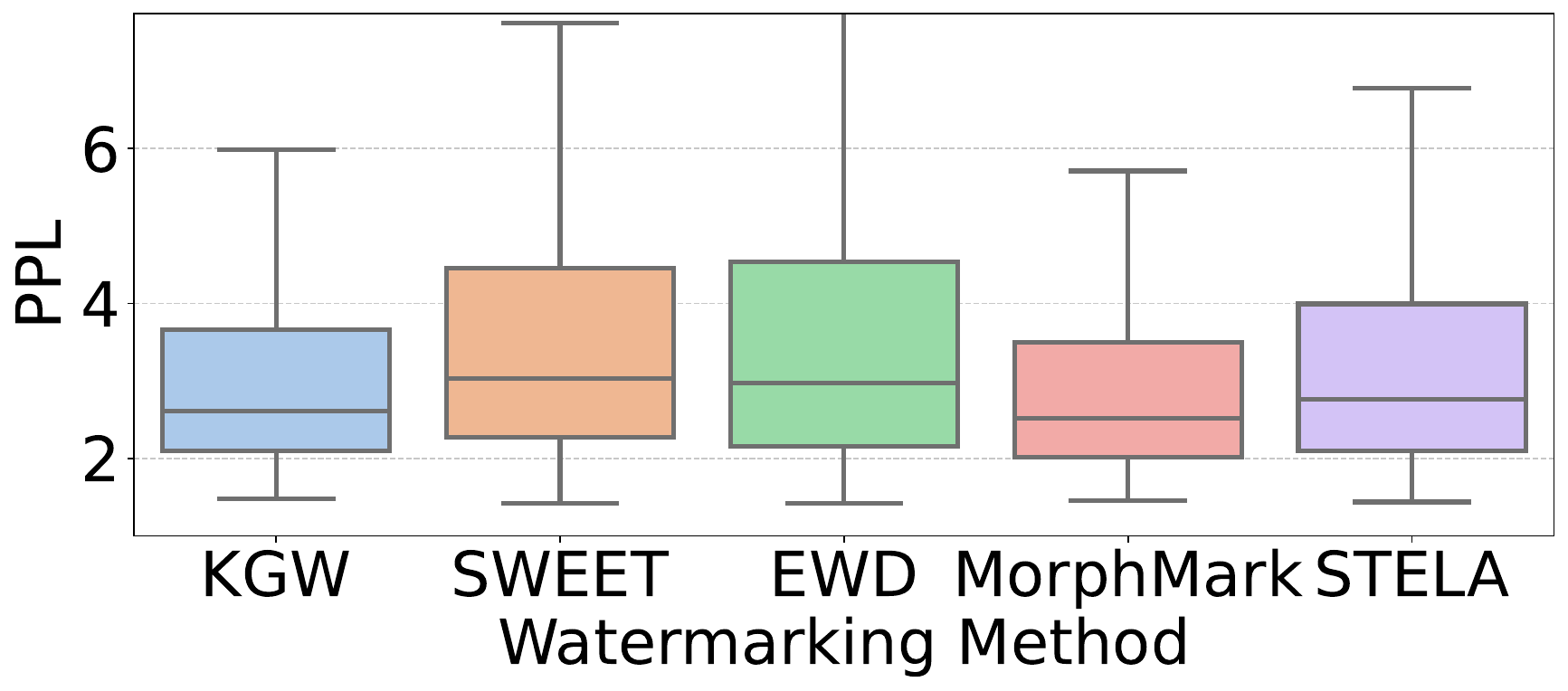}
    \end{subfigure}\hfill
    \begin{subfigure}{0.32\textwidth}
        \includegraphics[width=\linewidth]{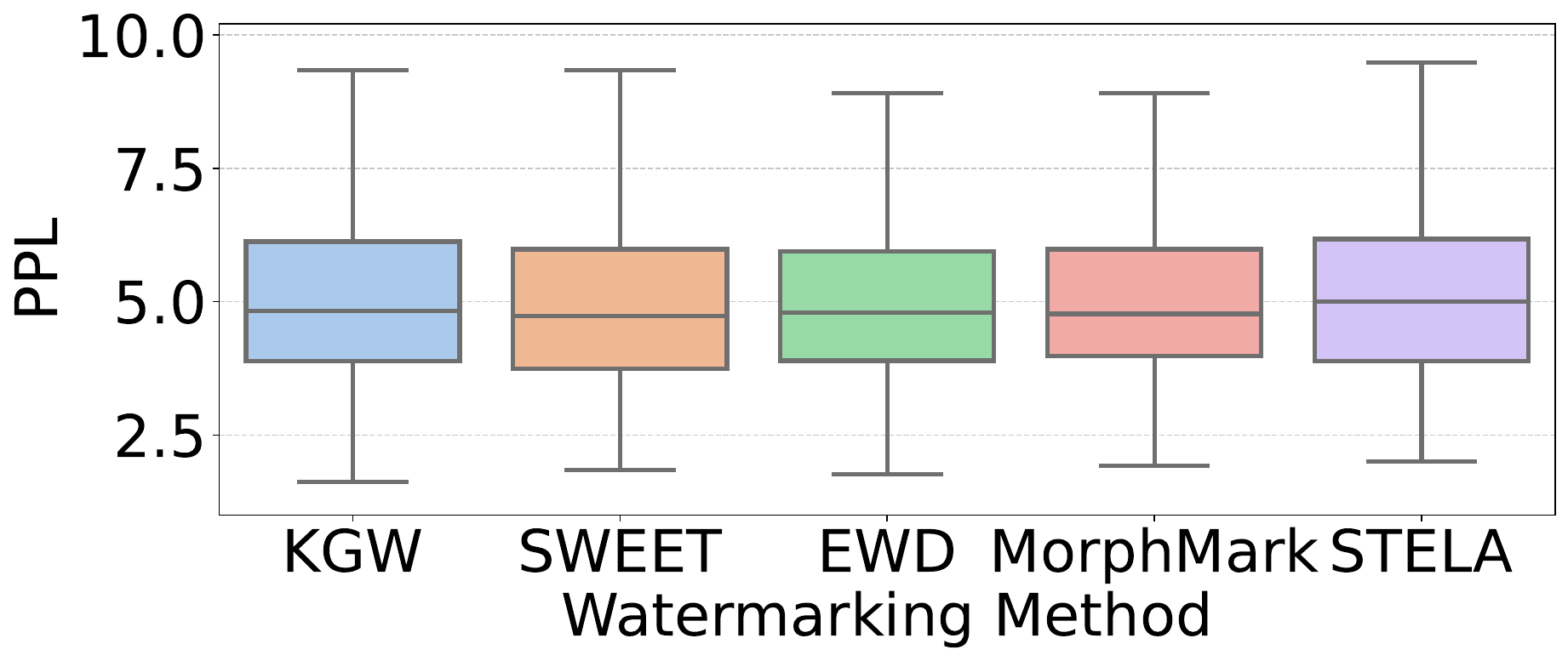}
    \end{subfigure}\hfill
    \begin{subfigure}{0.32\textwidth}
        \includegraphics[width=\linewidth]{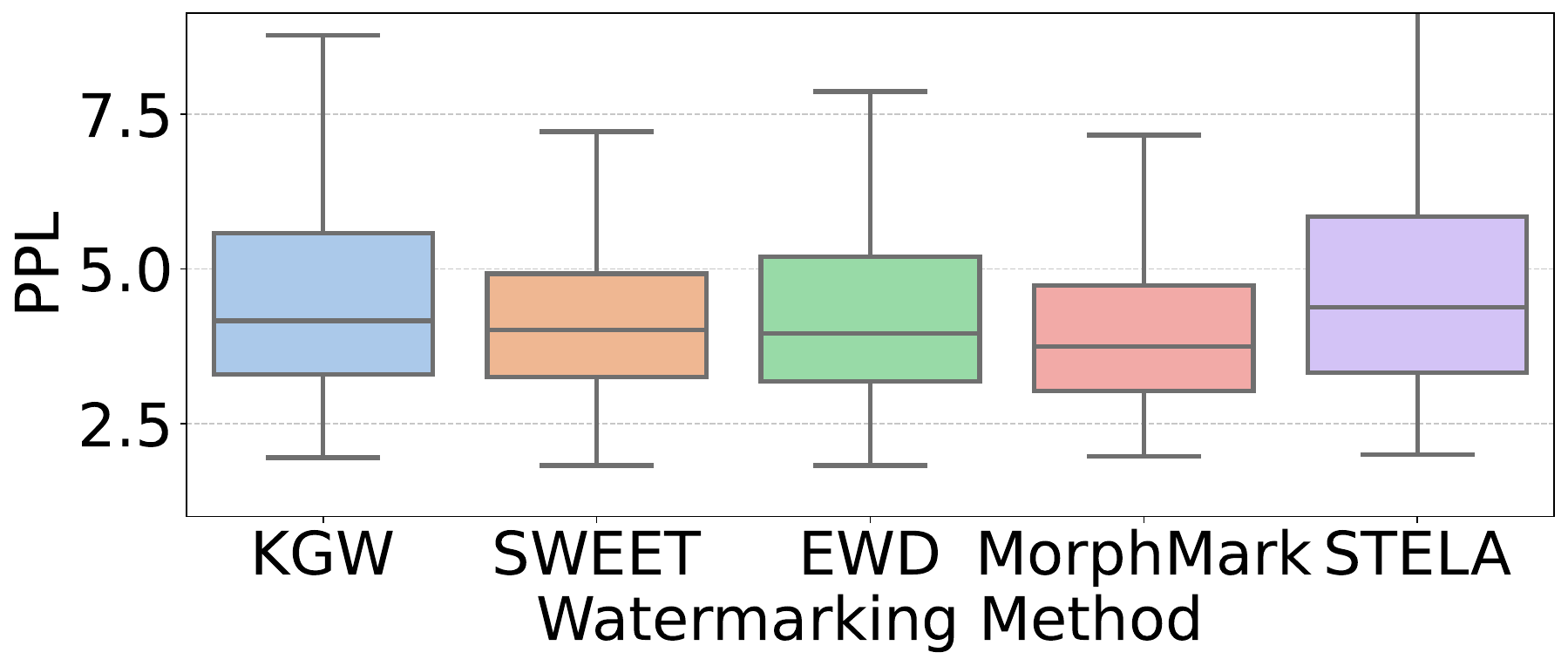}
    \end{subfigure}

    \begin{subfigure}{0.32\textwidth}
        \includegraphics[width=\linewidth]{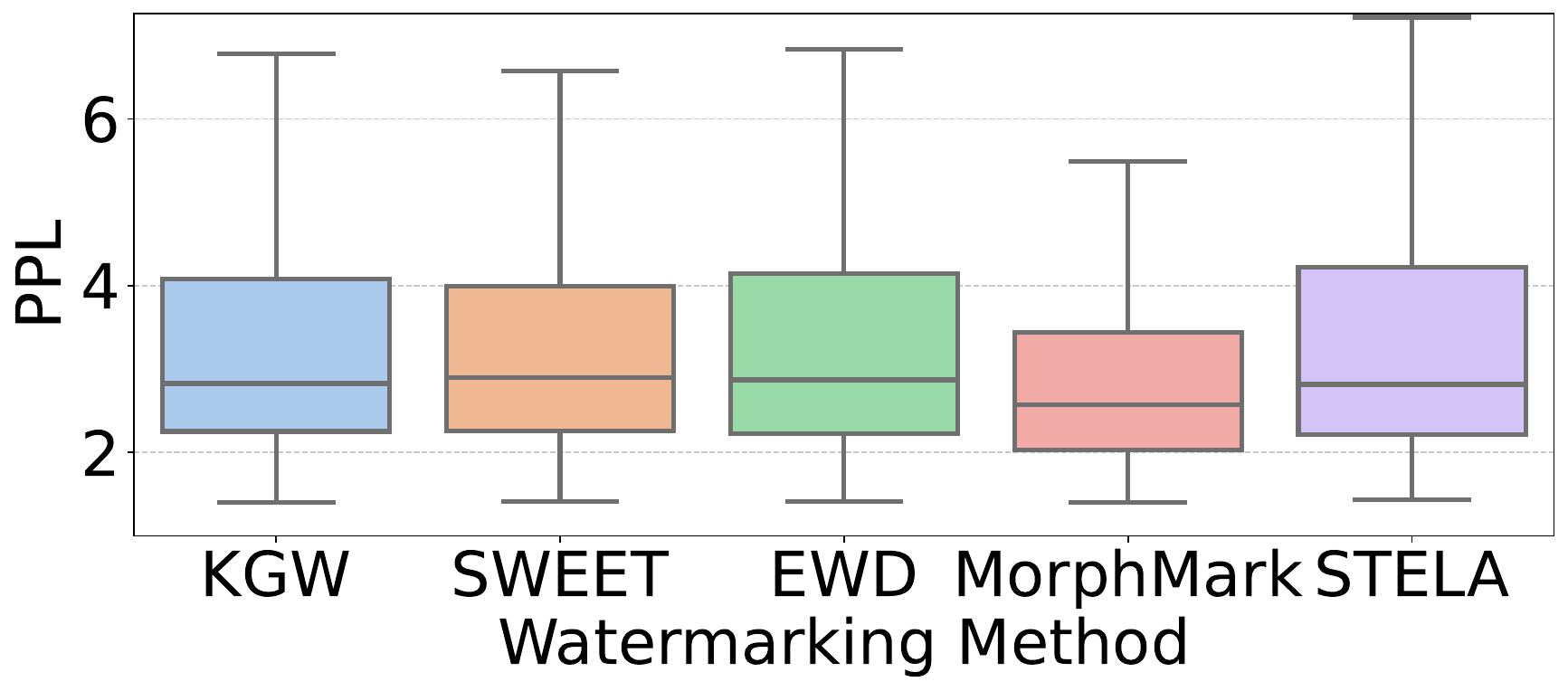}
    \end{subfigure}\hfill
    \begin{subfigure}{0.32\textwidth}
        \includegraphics[width=\linewidth]{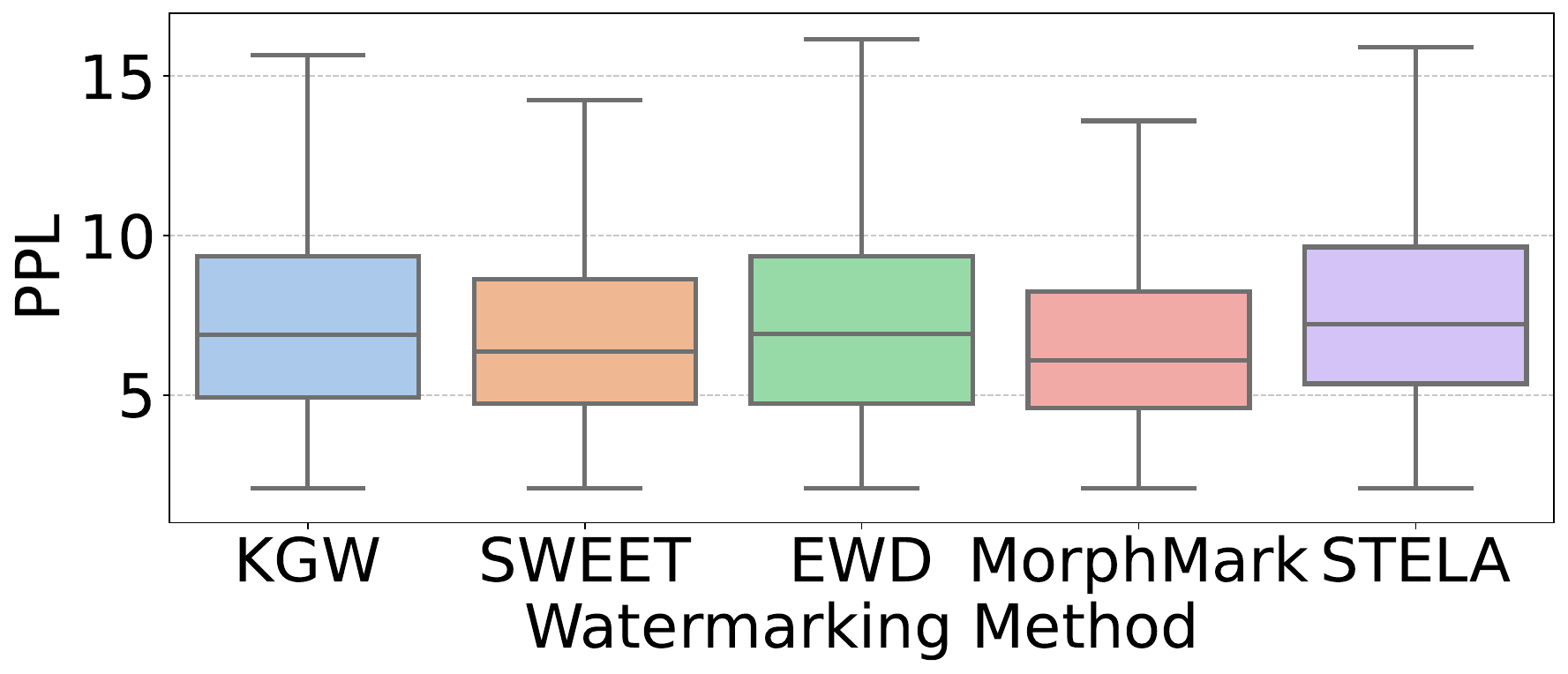}
    \end{subfigure}\hfill
    \begin{subfigure}{0.32\textwidth}
        \includegraphics[width=\linewidth]{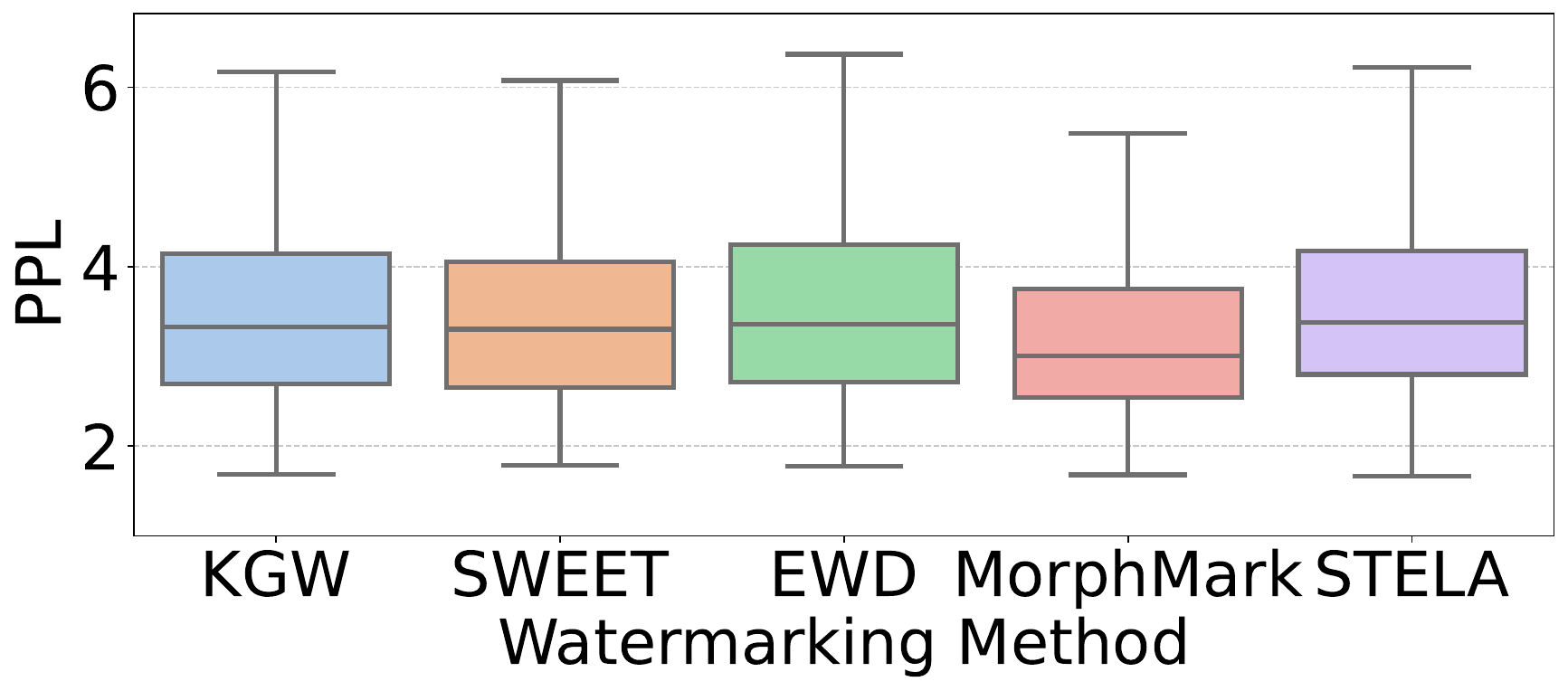}
    \end{subfigure}

    \caption{Perplexity of watermarked texts. 
    Columns represent languages~(English, Chinese, Korean) and 
    rows represent LLMs~(Llama-3.2, Qwen-3, HyperCLOVA).}
    \label{fig:ppl_results}
\end{figure*}

\subsection{Implementation Details}
\label{sec:implementation}

For reproducibility, 
we implement all baseline methods using the open-source 
MarkLLM toolkit~\citep{pan-etal-2024-markllm}. 
We set the green list ratio $\gamma=0.5$ for all methods, 
following the default configuration of the toolkit. 
For the baselines, we use the recommended 
watermark strength~($\delta$): 2.0 for KGW, EWD, and SWEET, 
and 1.3 for MorphMark. 
We set the generation temperature to 0.7.

To ensure a fair comparison, 
we calibrate the average watermark strength 
to be consistent across methods.
A naive application of a fixed $\delta$ would 
place STELA at an unfair disadvantage or advantage 
depending on the average linguistic indeterminacy 
of the language. 
Therefore, we calibrate the base strength for STELA in each language
to match the effective strength of the primary baselines. 
Specifically, we set $\delta = 2.0 / \mathbb{E}[\lambda(c_t)]$, 
where $\mathbb{E}[\lambda(c_t)]$ is the mean linguistic indeterminacy
over all POS contexts for that language. 
This calibration helps STELA maintain, on average, 
a comparable level of bias in the logit distribution to that of KGW, EWD, 
and SWEET. 
The estimated values of $\mathbb{E}[\lambda(c_t)]$ for English, Chinese, and Korean 
are 0.575, 0.523, and 0.475, respectively.
In Section~\ref{sec:pos_tagset_granularity}, we compare the use of language-specific 
versus universal POS tagsets, analyzing both the difference in their 
estimated $\mathbb{E}[\lambda(c_t)]$ values and the consequent effect on performance.

\section{Results and Analysis}
\label{sec:results_analysis}

\subsection{Experimental Results}

\paragraph{Detection Performance.}
Table~\ref{tab:detection_results} shows that STELA consistently achieves 
the highest average detection performance, calculated across 
all three languages, for each of the three LLMs. 
The superiority of STELA is particularly evident in the Chinese 
and Korean languages, where it secures the top TPR@5\%FPR and Best F1 scores 
across all tested LLMs. 
This indicates that our linguistics-aware approach is especially effective 
in languages with complex syntactic structures that 
a simple static approach might miss.
Furthermore, when paired with HyperCLOVA, 
STELA outperforms all baselines in all three languages, 
highlighting its robust compatibility with diverse model architectures. 
Even in English, where the performance margin is narrower, 
STELA remains highly competitive with the baselines, 
confirming its overall effectiveness.

\paragraph{Text Quality.}
While achieving superior detection, STELA maintains a high level of 
text quality, comparable to that of leading baselines. 
Figure~\ref{fig:ppl_results} presents the perplexity distributions 
for watermarked texts. 
The PPL distributions for STELA appear statistically comparable to those 
of baselines. 
These results indicate that the significant gains in detection performance 
do not come at the cost of a substantial degradation in text quality. 
The adaptive mechanism, guided by linguistic indeterminacy, 
effectively mitigates the risk of generating unnatural text 
in grammatically constrained contexts. 
We further complement perplexity with a more holistic signal 
via an LLM-as-Judge pairwise blind evaluation~(A/B test), 
in which STELA is preferred or tied over KGW in 52.5\% of 1,000 
comparisons, corroborating the perplexity parity reported 
above. 
Full experimental details are provided in Appendix~\ref{appendix:ab_test}.

\paragraph{Cross-Lingual and Cross-Model Generalization.}
The consistently high performance of STELA across analytic English, 
isolating Chinese, and agglutinative Korean validates our hypothesis 
that its mechanism is grounded in a language-universal principle of 
syntactic predictability, rather than the idiosyncrasies of a single grammar.
Furthermore, the robust performance of STELA across three different LLMs, 
each with distinct architectures and training philosophies, 
demonstrates the generalizability of our model-independent approach. 
The linguistic signal provides a stable foundation for watermarking that 
is not sensitive to the internal variations of specific models. 

\subsection{Ablation Study on POS Context Length}
\label{sec:pos_context_length}

We investigate how the optimal syntactic context length, $k$, 
varies according to a language's typological characteristics. 
Our hypothesis is that a shorter context~(e.g., $k=2$)
is sufficient for a language with a relatively fixed word order like English. 
In contrast, we anticipate that a wider context~(e.g., $k=4$)
is necessary for languages like Korean and Chinese 
to better capture longer-range syntactic dependencies.

\begin{figure}[hbt!]
    \centering
        \includegraphics[width=\columnwidth]{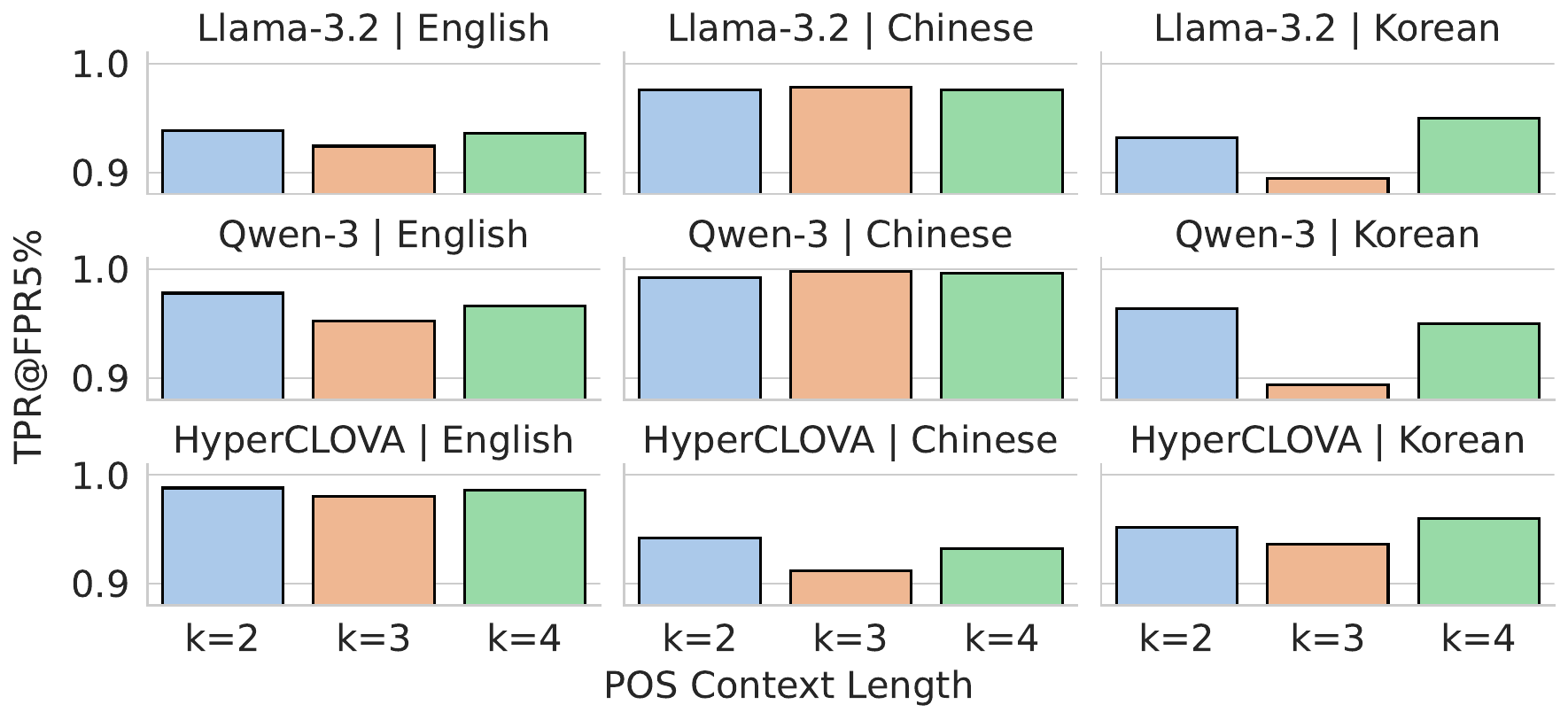}
    \caption{The effect of POS context length $k$ on the watermark detection performance of 
    STELA. 
    Each subplot corresponds to a specific LLM~(rows) and language~(columns). 
    }\label{fig:pos_context_length}
\end{figure}

The results in Figure~\ref{fig:pos_context_length} support this hypothesis.
In English, performance is near-optimal at $k=2$ and remains stable. 
This outcome is attributed to its relatively fixed SVO word order, 
where the immediately preceding POS tag provides 
a strong predictive signal for the next token's grammatical role.
Conversely, both Korean and Chinese show improved performance with a wider context of $k=4$. 
This suggests a larger window is needed to model dependencies in Korean's flexible, 
case-marked word order, as well as the more complex phrasal structures found in Chinese.

\subsection{Robustness to POS Tagset Granularity}
\label{sec:pos_tagset_granularity}

To validate STELA's practical utility, we evaluate its robustness to the granularity of its core input signal: the POS tagset.
We analyze the performance variation of STELA when using two distinct tagsets 
that offer different levels of grammatical detail: 
a cross-lingually consistent universal dependencies~(UD) tagset 
and fine-grained, language-specific tagsets. 
We apply the UD tagset using the spaCy library, employing
\texttt{en\_core\_web\_sm}, \texttt{zh\_core\_web\_sm}, and
\texttt{ko\_core\_news\_sm} for English, Chinese, and Korean, respectively.

Table~\ref{tab:pos_tagset_results} shows STELA performs well with both universal and 
language-specific POS tagsets. With the simpler universal tagset, 
STELA achieves high detection rates across all three languages, 
indicating it does not require fine-grained grammatical annotation. 
Language-specific tagsets further improve performance, 
especially for Korean, suggesting STELA benefits from 
more detailed grammatical distinctions when available.

\begin{table*}[hbt!]
\centering\small
\begin{tabular}{llc|ccc}
\toprule
\multirow{2}{*}{Language} & \multirow{2}{*}{POS Tagset} & \multirow{2}{*}{$\mathbb{E}[\lambda(c_t)]$} & Llama-3.2 & Qwen-3 & HyperCLOVA \\
\cmidrule(lr){4-4} \cmidrule(lr){5-5} \cmidrule(lr){6-6}
& & & TPR@FPR5\% & TPR@FPR5\% & TPR@FPR5\% \\
\midrule

\multirow{2}{*}{English}
& Universal & 0.686 & \textbf{0.948} & 0.972 & 0.984 \\
& Language-specific & 0.575 & 0.938 & \textbf{0.978} & \textbf{0.988} \\
\cmidrule(lr){1-6}

\multirow{2}{*}{Chinese}
& Universal & 0.643 & \textbf{0.976} & \textbf{0.998} & 0.930 \\
& Language-specific & 0.523 & \textbf{0.976} & 0.996 & \textbf{0.932} \\
\cmidrule(lr){1-6}

\multirow{2}{*}{Korean}
& Universal & 0.647 & 0.928 & 0.932 & 0.950 \\
& Language-specific & 0.475 & \textbf{0.950} & \textbf{0.950} & \textbf{0.960} \\

\bottomrule
\end{tabular}
\caption{
Watermark detection performance of STELA, using universal versus language-specific 
POS tagsets across three languages. 
We conduct this analysis by setting $k=2$ for English and 
$k=4$ for both Chinese and Korean.
We report the average linguistic indeterminacy~($\mathbb{E}[\lambda(c_t)]$) estimated 
by each tagset, alongside the resulting detection performance for three LLMs. 
The best performance for each model-language pair is highlighted in bold.
}
\label{tab:pos_tagset_results}
\end{table*}

An interesting observation is that 
the average linguistic indeterminacy~($\mathbb{E}[\lambda(c_t)]$) is lower for the 
language-specific tagsets than for the universal one. 
We attribute this to the ability of a finer-grained tagset to better distinguish between syntactically 
constrained positions~(i.e., $\lambda(c_t) \rightarrow 0$) and positions with greater 
freedom~(i.e., $\lambda(c_t) \rightarrow 1$). 
For example, while the UD tagset might classify all Korean particles as a single PART tag, 
a language-specific tagset distinguishes between nominative~(JKS) and objective~(JKO) 
case markers. 
This fine-grained distinction provides a more precise signal of grammatical constraint, 
enabling more effective watermark allocation.
In conclusion, these results suggest STELA adapts  
its performance with the level of available grammatical information.

\subsection{Decomposing the Watermark Signal}

We analyze the contribution of different word categories to the detection z-score 
to understand how STELA distributes watermarks.
For a fair cross-linguistic comparison, 
we employ the universal POS tagset for all three languages.
Following the universal dependencies standard\footnote{\url{https://universaldependencies.org/u/pos/}}, 
we classify tokens as content words~(open-class), 
function words~(closed-class), and others. 
A detailed description of the UD tags is provided in Appendix~\ref{appendix:upos}.
The tags are grouped as follows:
\begin{itemize}
    \item Content Words: ADJ, ADV, INTJ, NOUN, PROPN, VERB.
    \item Function Words: ADP, AUX, CCONJ, DET, NUM, PART, PRON, SCONJ.
    \item Others: PUNCT, SYM, X.
\end{itemize}
The analysis estimates the individual contribution of each token to the 
z-score numerator and subsequently aggregates these estimates across 
the designated linguistic categories.

\begin{figure}[hbt!]
    \centering
    \begin{subfigure}[b]{0.32\columnwidth}
        \centering
        \includegraphics[width=\linewidth]{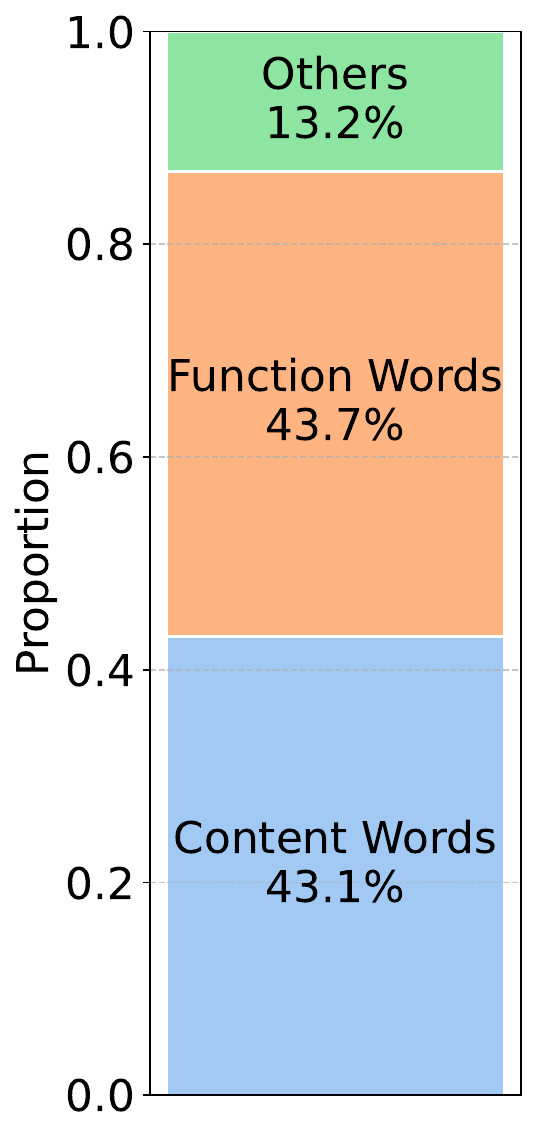}
        \caption{English}
        \label{fig:en_z_score}
    \end{subfigure}
    \hfill 
    \begin{subfigure}[b]{0.32\columnwidth}
        \centering
        \includegraphics[width=\linewidth]{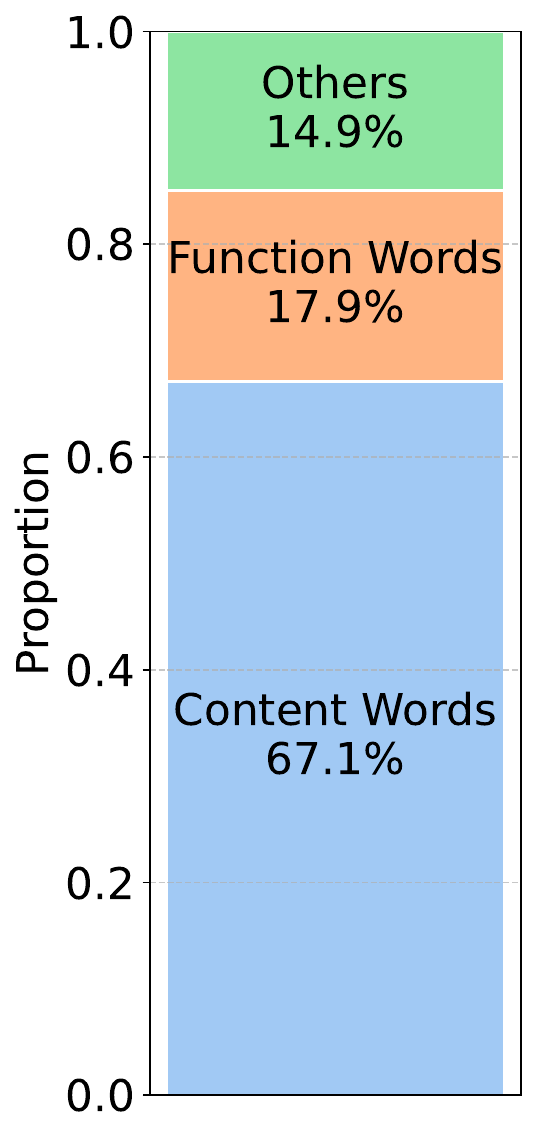}
        \caption{Chinese}
        \label{fig:zh_z_score}
    \end{subfigure}
    \hfill 
    \begin{subfigure}[b]{0.32\columnwidth}
        \centering
        \includegraphics[width=\linewidth]{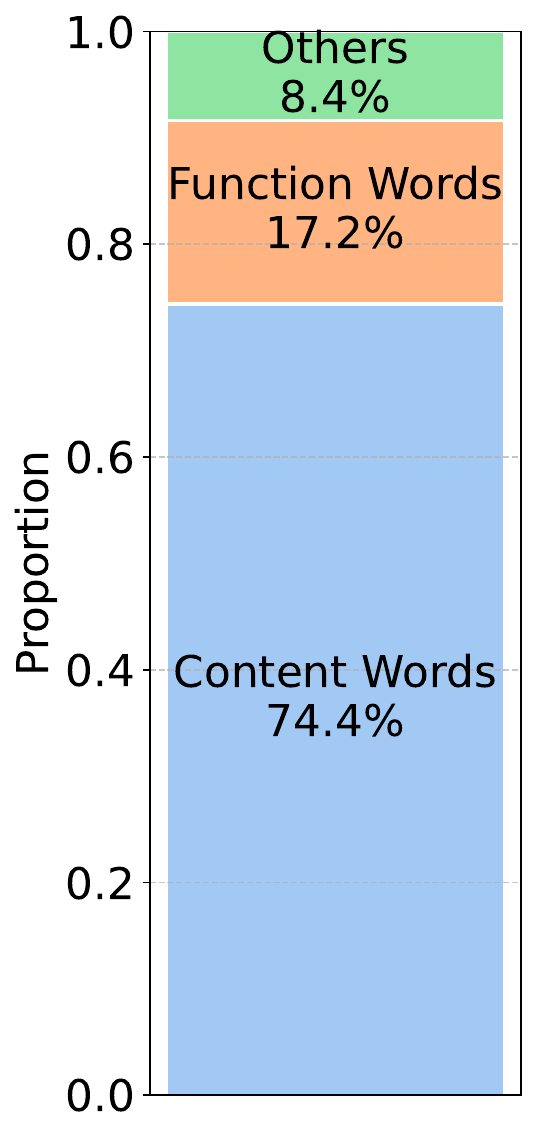}
        \caption{Korean}
        \label{fig:ko_z_score}
    \end{subfigure}
    \caption{Proportional contribution of word categories to the total watermark detection 
    z-score from STELA.}
    \label{fig:z_score_analysis}
\end{figure}

Figure~\ref{fig:z_score_analysis} 
visualizes the proportional contribution of each category to the total z-score.
The analysis presented here focuses on a representative model for each language: 
Llama-3.2 for English, Qwen-3 for Chinese, and HyperCLOVA for Korean. 
See Appendix~\ref{appendix:decomposing_z_score} for all model-language pairs.
The results show a strong correspondence between the watermark distribution 
and the typological characteristics of each language. 
In English, content words~(43.1\%) and function words~(43.7\%) 
contribute almost equally to the z-score. 
This reflects a core feature of analytic languages, where function words like prepositions 
and determiners are central to the grammatical structure. 
STELA thus identifies the contexts preceding these function words as having high 
linguistic freedom and allocates the watermark signal accordingly.

In contrast, the contribution of content words is dominant in Chinese~(67.1\%) 
and even more pronounced in Korean~(74.4\%).
This observation is consistent with the typological characteristics of these 
languages, a point further elaborated 
in Appendix~\ref{appendix:linguistic_background}.
In an agglutinative language such as Korean, 
grammatical functions are expressed through morphemes attached directly to content word stems.
The selection of the content word stem itself therefore provides considerable freedom, 
which STELA regards as an appropriate site for watermark insertion.
This analysis thus confirms that the watermark's distribution directly reflects the typological characteristics of each language.

\subsection{Robustness against Adversarial Attacks}
\label{adversarial_attacks}

\begin{figure}[hbt!]
    \centering
        \includegraphics[width=0.9\columnwidth]{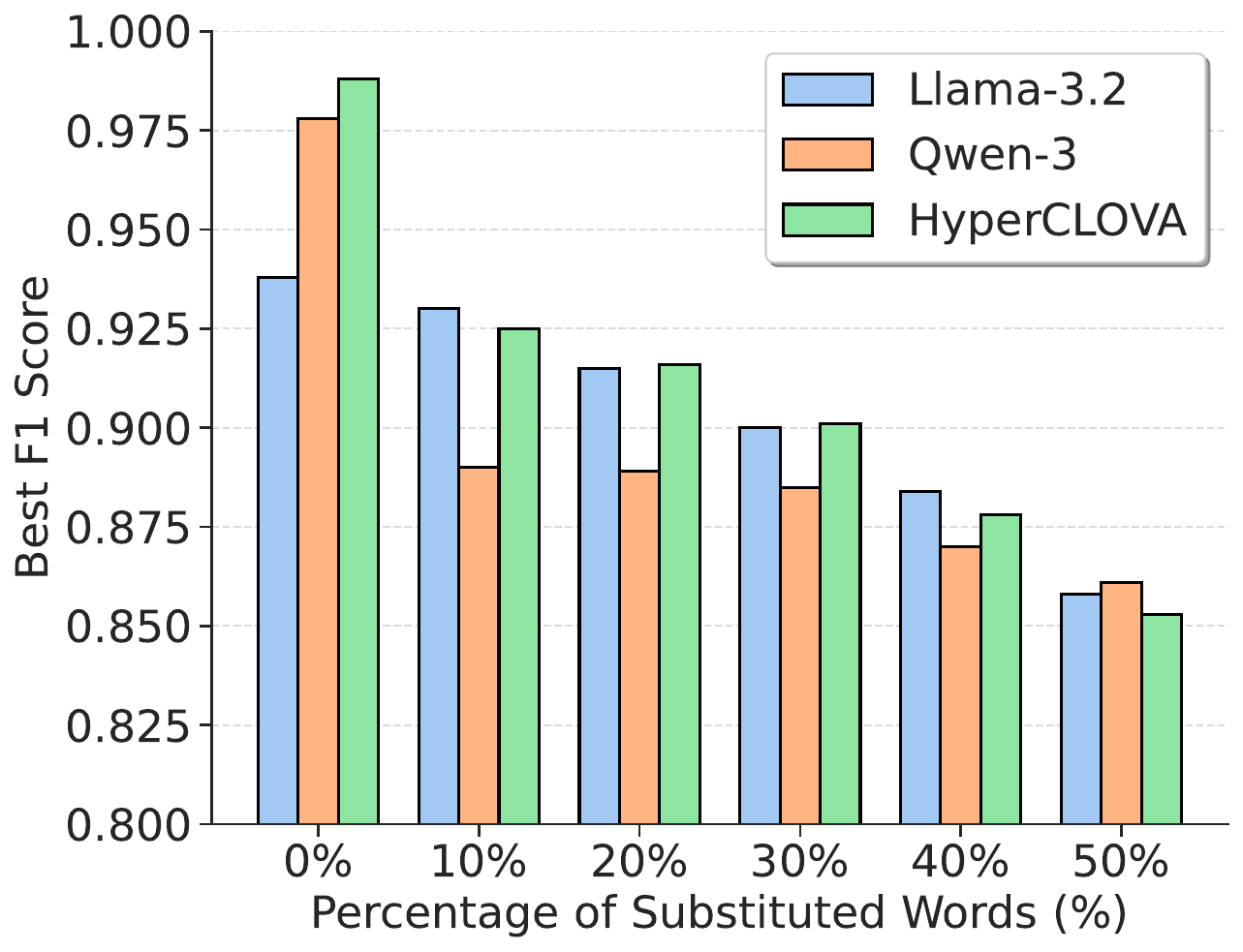}
    \caption{
    The plot compares the Best F1 Score for watermark detection across 
    three models as the percentage of words replaced with WordNet 
    synonyms increases.
    Notably, even when 50\% of the words in the text are replaced with synonyms, 
    all models maintain a high watermark detection F1 score of over 0.85. 
    }\label{fig:adversarial_attack}
\end{figure} 

We evaluate the resilience of STELA against an adversarial attack simulation 
based on synonym substitution. 
This experiment measures the persistence of the watermark signal 
when a malicious actor attempts its removal via paraphrasing. 
For this analysis, we subject watermarked English texts to a 
synonym replacement attack of varying intensity, sourcing synonyms 
from WordNet~\citep{fellbaum1998wordnet}. 
The attack intensity is controlled by adjusting the word substitution rate, 
which varies from 10\% to 50\%. 
We measure the watermark detection performance using the Best F1 Score.

Figure~\ref{fig:adversarial_attack} presents the results 
across the three tested LLMs. 
The findings demonstrate the robustness of STELA. 
Even under a substantial attack where 50\% of the words in the text 
are replaced, the detection F1 score remains high, consistently above 0.85 
for all models. 
This outcome indicates that the linguistic signal leveraged by STELA 
is deeply embedded within the syntactic structure of the text and is not 
easily erased through simple lexical alterations. 
The strength of the watermark is distributed across various parts of the text, 
which renders the signal resilient to localized changes.

A complementary and more challenging question is whether STELA 
also withstands structural rewriting that modifies 
phrasing at the clause and sentence level. 
We address this question through an additional analysis using the 
Dipper paraphrase attacker~\cite{krishna2023paraphrase}, 
a state-of-the-art neural paraphraser known to evade many AI-text detectors.

We focus on a single representative configuration in order 
to isolate the effect of paraphrasing intensity: 
English texts generated by Llama-3.2, drawn from the same 
evaluation set used in our main experiments. 
Watermarked texts are passed through Dipper, 
and we vary the lexical diversity parameter $L \in \{10, 20, 40, 50\}$, 
representing increasing degrees of paraphrasing intensity from light to heavy. 
We report the Best F1 score for watermark detection under each setting.

As shown in Table~\ref{tab:dipper}, the results demonstrate 
remarkable resilience across all attack intensities. 
Even under standard paraphrasing~($L{=}40$), which 
introduces substantial lexical and structural changes, STELA 
maintains a Best F1 score of 0.813. 
Notably, detection performance does not collapse under heavy attack~($L{=}50$), with the F1 score remaining robust at 0.825.

\begin{table}[hbt!]
\centering
\small
\begin{tabular}{l|c}
\toprule
\textbf{Attack Intensity} & \textbf{Best F1} \\
\midrule
Baseline (No Attack)     & 0.953 \\
\midrule
$L{=}10$ (Light)          & 0.862 \\
$L{=}20$ (Moderate)       & 0.869 \\
$L{=}40$ (Standard)       & 0.813 \\
$L{=}50$ (Heavy)          & 0.825 \\
\bottomrule
\end{tabular}
\caption{Detection performance of STELA under Dipper 
paraphrase attacks on English texts generated by Llama-3.2. 
Even under heavy paraphrasing~($L{=}50$), STELA maintains a 
Best F1 score above 0.80.}
\label{tab:dipper}
\end{table}

\section{Conclusion}

We introduce STELA, a watermarking method that uses syntactic predictability 
to balance text quality with detection robustness. 
By modulating strength based on a model-free linguistic signal, 
STELA achieves superior detection across 
typologically diverse languages. 
This is achieved by weakening the signal in grammatically constrained contexts 
to preserve quality, while strengthening it where linguistic choices 
are abundant to enhance detectability. 
This linguistics-aware framework opens new horizons for developing governance 
tools that are not only effective but also interpretable, 
harmonizing advanced AI systems with the foundational principles of human language.

\section*{Limitations}

While STELA exhibits strong performance and broad generalizability, 
several limitations remain that point to directions for future investigation.

First, the effectiveness of the method depends on the reliability of the 
part-of-speech tagger. 
Tagging errors during detection may distort contextual reconstruction 
and lead to missed watermark signals. 
Future work could explore methods to enhance robustness against 
tagging inaccuracies.

Second, our assessment of text quality currently relies on perplexity. 
While perplexity serves as a standard metric for fluency, it may not fully capture 
other critical dimensions of quality, such as grammatical correctness, 
stylistic naturalness, or semantic coherence. 
Developing more comprehensive evaluation metrics, specifically designed to assess 
the nuanced impact of watermarking from these multiple perspectives, 
remains an important area for future research.

Third, the linguistic indeterminacy signal is estimated from reference corpora. 
When the domain of these corpora diverges substantially from that of the generated text, 
the resulting weighting scheme may become suboptimal. 
Developing domain-adaptive models of syntactic predictability 
could further extend the applicability of the approach across diverse text sources.

Finally, while this study covers three major 
language types~(analytic, isolating, and agglutinative), 
it does not encompass the full typological diversity of the world’s languages, 
such as fusional or polysynthetic systems. 
Moreover, its application in low-resource settings presents a pressing challenge, 
given its dependency on high-quality POS taggers. 
Although our results with the universal dependencies tagset 
suggest a potential path for applying STELA to such languages, 
its effectiveness may be diminished where the performance of the tagger itself 
is inherently low. 
A thorough investigation into its performance under these constraints 
is therefore imperative for establishing its equitable and widespread applicability.

\section*{Acknowledgments}
This research was supported by the NRF grant~(RS-2025-00562134) and the AI Graduate School Program~(RS-2020-II201361) funded by the Korean government.


\bibliography{custom}

\appendix

\section{Further Discussion of Related Work}
\label{appendix:related_work}

This section provides a detailed overview of recent advancements in LLM watermarking. 
We group these methods by their core technical contributions.

\paragraph{Partitioning-based Watermarking.}
A substantial line of research builds upon the KGW framework by addressing 
two critical challenges inherent in static vocabulary partitioning: 
the degradation of text quality and the susceptibility to paraphrasing attacks.
These studies primarily seek to incorporate semantic and lexical information 
to design more adaptive and context-aware partitioning strategies.

\citet{guo2024context}
propose a semantic-aware watermark to enhance 
both robustness against paraphrasing attacks and text quality. 
Their method has two core components. 
First, it generates a watermark key by applying locality-sensitive hashing~(LSH) 
to the semantic embedding of the preceding context. 
As LSH tends to produce the same hash value for semantically similar inputs, 
the watermark key is less likely to change even if an attacker replaces words with synonyms, 
thus improving resilience. 
Second, instead of randomly partitioning the entire vocabulary, 
their method first uses LSH to group the vocabulary into sets of semantically similar 
tokens and then partitions the green/red lists within each semantic set. 
This semantically balanced partitioning scheme prevents a scenario where all words 
expressing a certain meaning are assigned to the red list, thereby preserving text quality.

\citet{chen-etal-2024-watme} also address the issue of text quality degradation 
caused by the unavailability of suitable tokens under arbitrary vocabulary partitioning.
They propose a method that leverages lexical redundancy.
The approach first identifies clusters of interchangeable tokens~(e.g., synonym sets) 
within the vocabulary using external dictionaries or an LLM.
During the green/red list partitioning, a mutual exclusion rule is applied to these clusters,
which enables the inclusion of remaining synonyms in the green list 
even when some for a given meaning are allocated to the red list.

\paragraph{Sampling-based Watermarking.}
Sampling-based watermarks operate 
by altering the token selection process itself. 
These approaches aim to embed a signal without changing the underlying 
probability distribution.

\citet{kuditipudi2024robust} 
propose a distortion-free watermarking methodology based on this principle. 
Their core idea is to deterministically map a sequence of random numbers, 
generated from a secret key, to a sequence of tokens sampled from the language model. 
They introduce two specific sampling schemes to achieve this: 
one based on inverse transform sampling and another on exponential minimum sampling. 
The detection process does not require the original prompt or model; 
instead, it uses sequence alignment techniques to identify a correlation between a 
given text and the expected random number sequence generated by the key. 

\citet{dathathri2024scalable} introduce SynthID-Text, 
a production-ready watermarking scheme that also modifies the sampling procedure. 
Their novel sampling algorithm, tournament sampling, involves an over-generation and 
filtering process. 
At each step, multiple candidate tokens are sampled from the model's original distribution. 
These candidates then compete in a tournament, 
where the winner at each stage is selected based on pseudorandom scoring functions 
derived from the watermark key. 
The final winning token is chosen as the output. 
This method can be configured to be distortion-free or distortionary %to trade text quality 
enabling a trade-off between text quality
for higher detectability. 
Notably, SynthID-Text has been deployed at scale in Google Gemini.

\section{Algorithms of STELA}
\label{appendix:algorithms}

Algorithm~\ref{alg:estimation} details the 
pre-computation of the linguistic indeterminacy signal.
Algorithm~\ref{alg:insertion} describes the 
adaptive watermark insertion process during text generation.
Algorithm~\ref{alg:detection} outlines the procedure for watermark detection 
using adaptive scoring.

\begin{algorithm*}[hbt!]
\caption{Estimation of Linguistic Indeterminacy}
\label{alg:estimation}
\begin{algorithmic}[1]
\State \textbf{Require:} Reference human-written corpus $\mathcal{C}$, POS context size $k$
\State \textbf{Ensure:} Lookup table $\lambda$ mapping each POS context $c_t$ to its indeterminacy value

\State Initialize counts: $\text{count}(c) \gets 0$, $\text{count}(c, \pi) \gets 0$
\State Initialize POS tagger $\mathcal{T}$
\State $S_{\text{tagged}} \gets \mathcal{T}(\mathcal{C})$ \Comment{Tag the entire corpus with POS tags}

\Statex \Comment{Step 1: Count POS n-gram frequencies}
\For{each tagged sentence in $S_{\text{tagged}}$}
    \State $\Pi \gets \text{get\_pos\_tags}(\text{sentence})$
    \For{$t = k$ to $\text{length}(\Pi)$}
        \State $c_t \gets (\pi_{t-k+1}, \dots, \pi_{t-1})$ \Comment{Extract context of $k-1$ preceding POS tags}
        \State $\pi_t \gets \Pi[t]$
        \State $\text{count}(c_t) \gets \text{count}(c_t) + 1$
        \State $\text{count}(c_t, \pi_t) \gets \text{count}(c_t, \pi_t) + 1$
    \EndFor
\EndFor

\Statex \Comment{Step 2: Calculate linguistic indeterminacy $\lambda(c_t)$ for each context}
\State Initialize lookup table $\lambda$
\For{each unique context $c_t$ in counts}
    \State $H \gets 0$
    \State $\mathcal{V}_{\text{POS},c_t} \gets \{\pi \mid \text{count}(c_t, \pi) > 0\}$ \Comment{Set of observed POS tags following $c_t$}
    \For{each tag $\pi' \in \mathcal{V}_{\text{POS},c_t}$}
        \State $P(\pi'|c_t) \gets \frac{\text{count}(c_t, \pi')}{\text{count}(c_t)}$ \Comment{Estimate conditional probability}
        \State $H \gets H - P(\pi'|c_t) \log P(\pi'|c_t)$ \Comment{Conditional Shannon entropy (Eq.~\ref{eq:shannon_entropy})}
    \EndFor
    \State $K_{c_t} \gets |\mathcal{V}_{\text{POS},c_t}|$ \Comment{Number of unique tag types}
    \State $\lambda(c_t) \gets \frac{H}{\log K_{c_t}}$ \Comment{Normalized entropy (Eq.~\ref{eq:lambda})}
\EndFor
\State \Return{$\lambda$}
\end{algorithmic}
\end{algorithm*}

\begin{algorithm*}[hbt!]
\caption{Watermark Insertion with Adaptive Strength}
\label{alg:insertion}
\begin{algorithmic}[1]
\State \textbf{Require:} Preceding tokens $x_{<t}$, current logits $l_t$, base watermark strength $\delta$, green list ratio $\gamma$, POS tagger $\mathcal{T}$, pre-computed table $\lambda$, context size $k$
\State \textbf{Ensure:} Next watermarked token $x_t$

\Statex \Comment{Step 1: Determine the current linguistic context}
\State $\pi_{<t} \gets \mathcal{T}(x_{<t})$
\State $c_t \gets (\pi_{t-k+1}, \dots, \pi_{t-1})$ \Comment{Define the current POS context}

\Statex \Comment{Step 2: Compute adaptive watermark strength}
\State $\lambda_{\text{val}} \gets \lambda.\text{lookup}(c_t)$ \Comment{Retrieve linguistic indeterminacy value}
\State $\delta'_t \gets \delta \cdot \lambda_{\text{val}}$ \Comment{Modulate bias based on $\lambda(c_t)$ (Eq.~\ref{eq:adaptive_delta})}

\Statex \Comment{Step 3: Modify logits}
\State $\mathcal{V}_G \gets \Call{GenerateGreenList}{\text{hash}(x_{t-1}), \gamma}$ \Comment{Partition vocabulary}
\State $l'_t \gets l_t$
\For{$i = 1$ to $|\mathcal{V}|$}
    \If{$\text{token}_i \in \mathcal{V}_G$}
        \State $l'_{t,i} \gets l_{t,i} + \delta'_t$ \Comment{Apply adaptive bias to green-listed tokens (Eq.~\ref{eq:stela_logit_modification})}
    \EndIf
\EndFor

\Statex \Comment{Step 4: Sample the next token}
\State $p'_t \gets \text{softmax}(l'_t)$
\State $x_t \gets \Call{sample}{p'_t}$
\State \Return{$x_t$}
\end{algorithmic}
\end{algorithm*}

\begin{algorithm*}[hbt!]
\caption{Watermark Detection with Adaptive Scoring}
\label{alg:detection}
\begin{algorithmic}[1]
\State \textbf{Require:} Text sequence $X=(x_1, \dots, x_T)$, green list ratio $\gamma$, POS tagger $\mathcal{T}$, pre-computed table $\lambda$, context size $k$
\State \textbf{Ensure:} Weighted z-score $z'$

\Statex \Comment{Step 1: Initialization}
\State $W_G \gets 0$ \Comment{Sum of weights for green-listed tokens}
\State $\Sigma w \gets 0$ \Comment{Sum of all weights}
\State $\Sigma w^2 \gets 0$ \Comment{Sum of all squared weights}
\State $S_{\text{tagged}} \gets \mathcal{T}(X)$
\State $\Pi \gets \text{get\_pos\_tags}(S_{\text{tagged}})$

\Statex \Comment{Step 2: Iterate and accumulate weighted statistics}
\For{$t = k$ to $T$}
    \State $c_t \gets (\pi_{t-k+1}, \dots, \pi_{t-1})$ \Comment{Reconstruct POS context for $x_t$}
    \State $w_t \gets \lambda.\text{lookup}(c_t)$ \Comment{Token's weight is its linguistic indeterminacy (Eq.~\ref{eq:weight})}
    \State
    \State $\Sigma w \gets \Sigma w + w_t$
    \State $\Sigma w^2 \gets \Sigma w^2 + w_t^2$
    \State
    \State $\mathcal{V}_{G,t} \gets \Call{GenerateGreenList}{\text{hash}(x_{t-1}), \gamma}$ \Comment{Reconstruct green list}
    \If{$x_t \in \mathcal{V}_{G,t}$}
        \State $W_G \gets W_G + w_t$
    \EndIf
\EndFor

\Statex \Comment{Step 3: Calculate the final weighted z-score}
\State $\mu_{W_G} \gets \gamma \cdot \Sigma w$ \Comment{Expected value of $W_G$ under $H_0$}
\State $\sigma_{W_G} \gets \sqrt{\gamma(1-\gamma) \Sigma w^2}$ \Comment{Standard deviation of $W_G$ under $H_0$}
\State $z' \gets \frac{W_G - \mu_{W_G}}{\sigma_{W_G}}$ \Comment{Calculate final $z'$ score (Eq.~\ref{eq:weighted_z_score})}
\State \Return{$z'$}
\end{algorithmic}
\end{algorithm*}

\section{Language-Specific POS Tags}
\label{appendix:language_pos}

This section provides further details on the POS taggers selected 
for each language. 
These taggers are essential for deriving the linguistic indeterminacy 
signal central to our method. 
We select each tagger based on its performance, 
widespread adoption in the NLP community, 
and the ability of its tagset to capture the grammatical 
characteristics of each language. 
The tagsets for English, Korean, and Chinese are detailed in 
Tables~\ref{tab:en_tags},~\ref{tab:ko_tags}, and~\ref{tab:zh_tags}, respectively.

\paragraph{English: spaCy.}
For English, we utilize spaCy, a highly efficient NLP library. 
Its pre-trained models are widely recognized for their accuracy and robustness. 
We employ the model that uses the Penn Treebank tagset~(Table~\ref{tab:en_tags}), 
which provides a well-established and granular categorization of English 
parts-of-speech. 
This tagset effectively distinguishes between various 
function words~(e.g., determiners, prepositions) and content words, 
making it well-suited for modeling the syntactic structure of an 
analytic language like English.

\begin{table}[hbt!]
\centering
\small
\begin{tabular}{ll}
\hline
\textbf{Tag} & \textbf{Description} \\
\hline
CC & Coordinating conjunction \\
CD & Cardinal number \\
DT & Determiner \\
EX & Existential \textit{there} \\
FW & Foreign word \\
IN & Preposition or subordinating conjunction \\
JJ & Adjective \\
JJR & Adjective, comparative \\
JJS & Adjective, superlative \\
LS & List item marker \\
MD & Modal \\
NN & Noun, singular or mass \\
NNS & Noun, plural \\
NNP & Proper noun, singular \\
NNPS & Proper noun, plural \\
PDT & Predeterminer \\
POS & Possessive ending \\
PRP & Personal pronoun \\
PRP\$ & Possessive pronoun \\
RB & Adverb \\
RBR & Adverb, comparative \\
RBS & Adverb, superlative \\
RP & Particle \\
SYM & Symbol \\
TO & \textit{to} \\
UH & Interjection \\
VB & Verb, base form \\
VBD & Verb, past tense \\
VBG & Verb, gerund or present participle \\
VBN & Verb, past participle \\
VBP & Verb, non-3rd person singular present \\
VBZ & Verb, 3rd person singular present \\
WDT & Wh-determiner \\
WP & Wh-pronoun \\
WP\$ & Possessive wh-pronoun \\
WRB & Wh-adverb \\
. & Sentence-final punctuation (. ? !) \\
, & Comma \\
: & Colon or semicolon \\
-LRB-, -RRB- & Left/right round bracket \\
-LSB-, -RSB- & Left/right square bracket \\
\# & Pound sign \\
\$ & Dollar sign \\
\hline
\end{tabular}
\caption{Penn Treebank-based POS tags used in spaCy.}
\label{tab:en_tags}
\end{table}

%%%%%% 

\paragraph{Korean: Kiwi.}
Given the rich and complex morphology of Korean, 
we use Kiwi, a modern morphological analyzer specifically designed 
for the Korean language. 
Kiwi demonstrates excellent performance in accurately analyzing 
agglutinative structures, such as verb conjugations and case markers. 
It is based on an extended version of the Sejong POS tagset~(Table~\ref{tab:ko_tags}), 
a highly granular system that subdivides functional morphemes 
into detailed categories, including numerous types of particles~(\textit{josa}) 
and endings~(\textit{eomi}). 
This level of detail is critical for capturing the 
grammatical constraints in Korean.

\begin{table}[hbt!]
\centering
\small
\begin{tabular}{ll}
\hline
\textbf{Tag} & \textbf{Description} \\
\hline
NNG & General noun \\
NNP & Proper noun \\
NNB & Bound noun \\
NR  & Numeral \\
NP  & Pronoun \\
VV  & Verb \\
VA  & Adjective \\
VX  & Auxiliary predicate \\
VCP & Positive copula (\textit{이다}) \\
VCN & Negative copula (\textit{아니다}) \\
MM  & Determiner \\
MAG & General adverb \\
MAJ & Conjunctive adverb \\
IC  & Interjection \\
JKS & Subject particle \\
JKC & Complement (predicative) particle \\
JKG & Adnominal (genitive) particle \\
JKO & Object (accusative) particle \\
JKB & Adverbial particle \\
JKV & Vocative particle \\
JKQ & Quotation particle \\
JX  & Auxiliary particle \\
JC  & Conjunctive particle \\
EP  & Prefinal ending \\
EF  & Final ending \\
EC  & Conjunctive ending \\
ETN & Nominalizing ending \\
ETM & Adnominalizing ending \\
XPN & Noun prefix \\
XSN & Noun-derivational suffix \\
XSV & Verb-derivational suffix \\
XSA & Adjective-derivational suffix \\
XSM & Adverb-derivational suffix$^{*}$ \\
XR  & Root morpheme \\
SF  & Sentence-final punctuation (. ! ?) \\
SP  & Delimiter (, / : ;) \\
SS  & Quotation or parenthesis symbol \\
SSO & Opening quotation or bracket$^{*}$ \\
SSC & Closing quotation or bracket$^{*}$ \\
SE  & Ellipsis (\ldots) \\
SO  & Linking mark (hyphen, tilde) \\
SW  & Other special symbol \\
SL  & Latin alphabet (A--Z, a--z) \\
SH  & Chinese character (hanja) \\
SN  & Numeral (0--9) \\
SB  & Ordered list marker (e.g., 가., 나., 1., 2.)$^{*}$ \\
UN  & Unanalyzable token$^{*}$ \\
W\_URL    & URL address$^{*}$ \\
W\_EMAIL  & Email address$^{*}$ \\
W\_HASHTAG & Hashtag (\#abcd)$^{*}$ \\
W\_MENTION & Mention (@abcd)$^{*}$ \\
W\_SERIAL  & Serial number (phone/account/IP, etc.)$^{*}$ \\
W\_EMOJI   & Emoji$^{*}$ \\
Z\_CODA & Coda notation for Hangul$^{*}$ \\
Z\_SIOT & Syllable-final \textit{siot (ㅅ)} marker$^{*}$ \\
USER0\textasciitilde4 & User-defined tag$^{*}$ \\
\hline
\end{tabular}
\caption{Part-of-speech tags used in the Kiwi morphological analyzer, 
based on the Sejong POS system with extended and modified categories. 
Asterisks ($^{*}$) denote Kiwi-specific extensions beyond the Sejong POS set.}
\label{tab:ko_tags}
\end{table}

%%%%%%

\paragraph{Chinese: HanLP.}
For Chinese, we choose HanLP, a high-performance and popular NLP toolkit 
for the Chinese language. 
It offers robust support for the Penn Chinese Treebank standard, 
which we adopt for our analysis~(Table~\ref{tab:zh_tags}). 
This tagset is particularly effective for an isolating language like Chinese, 
as it includes fine-grained tags for crucial grammatical elements 
such as aspectual particles~(e.g., 了, 着), 
structural particles~(e.g., 的, 地, 得), and a wide range of measure words, 
which are essential for syntactic predictability.

\begin{table}[hbt!]
\centering
\small
\begin{tabular}{ll}
\hline
\textbf{Tag} & \textbf{Description} \\
\hline
AD & Adverb \\
AS & Aspect marker \\
BA & 把 in bar-construction \\
CC & Coordinating conjunction \\
CD & Cardinal number \\
CS & Subordinating conjunction \\
DEC & 的 in relative clause \\
DEG & Associative 的 \\
DER & 得 in V-de construction \\
DEV & 地 before VP \\
DT & Determiner \\
ETC & For words 等, 等等 \\
FW & Foreign word \\
IJ & Interjection \\
JJ & Other noun-modifier \\
LB & 被 in long 被-construction \\
LC & Localizer \\
M & Measure word \\
MSP & Other particle \\
NN & Common noun \\
NR & Proper noun \\
NT & Temporal noun \\
OD & Ordinal number \\
ON & Onomatopoeia \\
P & Preposition (excluding 被 and 把) \\
PN & Pronoun \\
PU & Punctuation \\
SB & 被 in short 被-construction \\
SP & Sentence-final particle \\
VA & Predicative adjective \\
VC & Copula “是” \\
VE & “有” as the main verb \\
VV & Other verb \\
\hline
\end{tabular}
\caption{Penn Chinese Treebank POS tags used in HanLP.}
\label{tab:zh_tags}
\end{table}

%%%%%

\begin{table}[hbt!]
\centering\small
\begin{tabular}{ll}
\hline
\textbf{Tag} & \textbf{Description} \\
\hline
ADJ & Adjective \\
ADP & Adposition \\
ADV & Adverb \\
AUX & Auxiliary \\
CCONJ & Coordinating Conjunction \\
DET & Determiner \\
INTJ & Interjection \\
NOUN & Noun \\
NUM & Numeral \\
PART & Particle \\
PRON & Pronoun \\
PROPN & Proper Noun \\
PUNCT & Punctuation \\
SCONJ & Subordinating Conjunction \\
SYM & Symbol \\
VERB & Verb \\
X & Other \\
\hline
\end{tabular}
\caption{Universal dependencies POS tags.}
\label{tab:ud_tags}
\end{table}

% New Camera-ready
\section{LLM-based Text Quality Assessment}
\label{appendix:ab_test}

We complement perplexity-based evaluation with a more holistic 
quality signal by conducting a 
pairwise blind evaluation~(A/B test) 
that compares STELA and KGW using GPT-5.1~(gpt-5.1-2025-11-13) as the judge. 
For 500 English prompts, we generate texts under both methods and instruct the 
evaluator to select the more natural, coherent, and 
grammatically correct continuation.
We conduct an A/B test using the following prompt.

\begin{quote}\small
\textit{Which of the following two texts is more natural, 
coherent, and grammatically correct given the preceding 
context? Focus on flow, word choice, and syntactic correctness.}

\textit{[Context]}\quad \textit{[Text A]} \quad \textit{[Text B]}

\textit{Select one: ``A is better'', ``B is better'', or 
``Tie''. Provide a brief reason.}
\end{quote}

We control for position bias by evaluating both 
presentation orderings~(STELA first and second), 
yielding 1,000 comparisons in total. 
As reported in Table~\ref{tab:ab_test}, 
the combined results show that STELA is preferred in 52.10\% 
of cases and preferred-or-tied in 52.50\% overall. 
This confirms that the adaptive mechanism produces text perceived 
as more natural than the static KGW baseline. 

\begin{table*}[hbt!]
\centering
\small
\begin{tabular}{lccc|c}
\toprule
\textbf{Scenario} & \textbf{STELA Wins} & \textbf{KGW Wins} 
  & \textbf{Tie} & \textbf{Win Rate~(STELA)} \\
\midrule
STELA First  & 179 & 319 & 2 & 35.80\% \\
STELA Second & 342 & 156 & 2 & 68.40\% \\
\midrule
\textbf{Combined} & \textbf{521} & \textbf{475} 
  & \textbf{4} & \textbf{52.10\%} \\
\bottomrule
\end{tabular}
\caption{GPT-5.1 pairwise blind evaluation on 
500 English prompts with 
bidirectional presentation~(1,000 total comparisons).}
\label{tab:ab_test}
\end{table*}

The substantial gap between 
Scenario~1~(STELA first: 35.8\% win rate) 
and Scenario~2~(STELA second: 68.4\% win rate), 
as shown in Table~\ref{tab:ab_test}, reveals a pronounced 
position bias in the evaluator, which systematically favors 
the second option presented. Our bidirectional evaluation 
design explicitly neutralizes this artifact by averaging 
over both orderings. The combined 52.1\% win rate therefore 
constitutes a bias-corrected estimate of relative quality 
preference, rather than an artifact of presentation order.

\section{Universal Dependencies POS Tags}
\label{appendix:upos}

Table~\ref{tab:ud_tags} presents the universal dependencies POS tags
and a description for each tag.

\begin{figure}[hbt!]
    \centering
    \begin{subfigure}[b]{0.32\columnwidth}
        \centering
        \includegraphics[width=\linewidth]{figures/z_score/k2_ud_english_llama-1B.pdf}
        \caption{Llama/En}
    \end{subfigure}
    \hfill 
    \begin{subfigure}[b]{0.32\columnwidth}
        \centering
        \includegraphics[width=\linewidth]{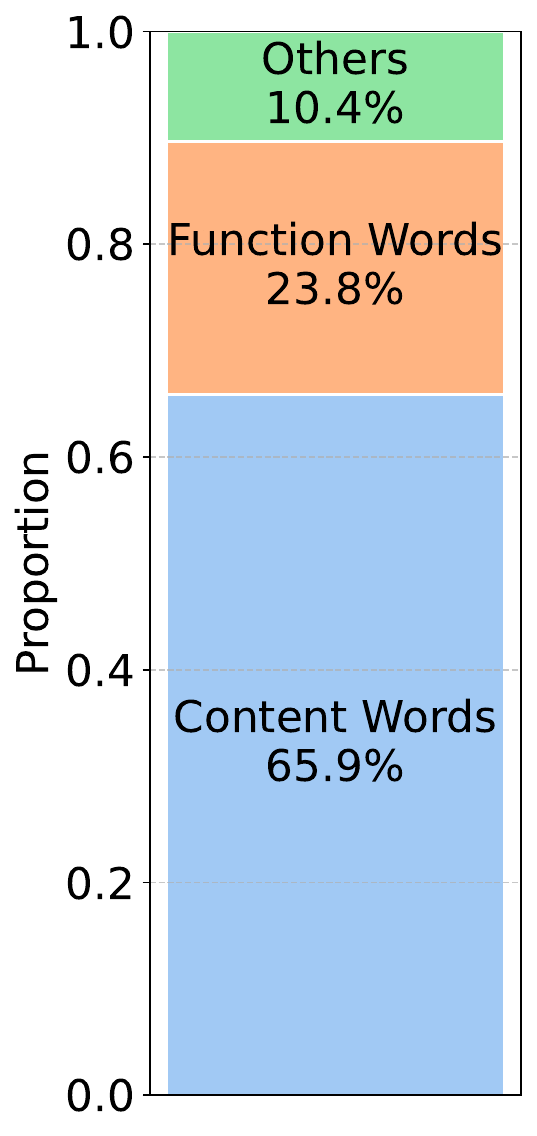}
        \caption{Llama/Zh}
    \end{subfigure}
    \hfill 
    \begin{subfigure}[b]{0.32\columnwidth}
        \centering
        \includegraphics[width=\linewidth]{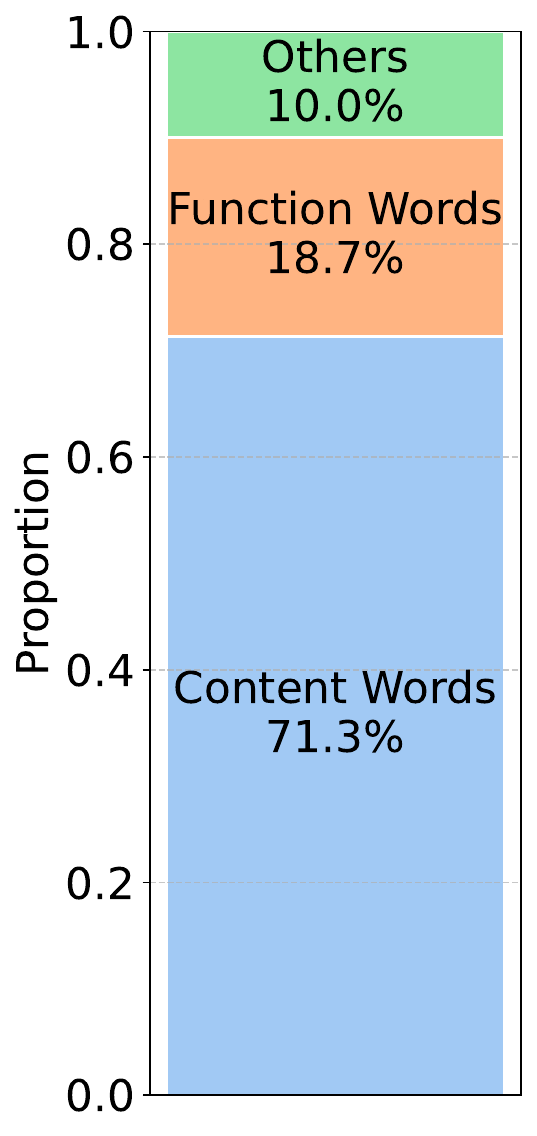}
        \caption{Llama/Ko}
    \end{subfigure}

    \vspace{1em} 

    \begin{subfigure}[b]{0.32\columnwidth}
        \centering
        \includegraphics[width=\linewidth]{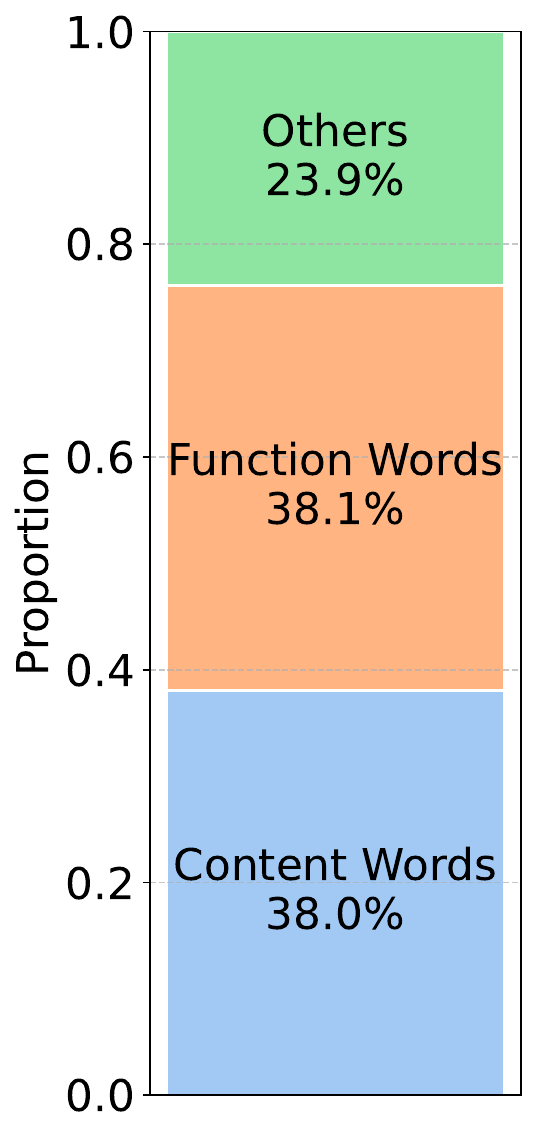} 
        \caption{Qwen/En}
    \end{subfigure}
    \hfill 
    \begin{subfigure}[b]{0.32\columnwidth}
        \centering
        \includegraphics[width=\linewidth]{figures/z_score/k4_ud_chinese_qwen3-0.6B.pdf} 
        \caption{Qwen/Zh}
    \end{subfigure}
    \hfill 
    \begin{subfigure}[b]{0.32\columnwidth}
        \centering
        \includegraphics[width=\linewidth]{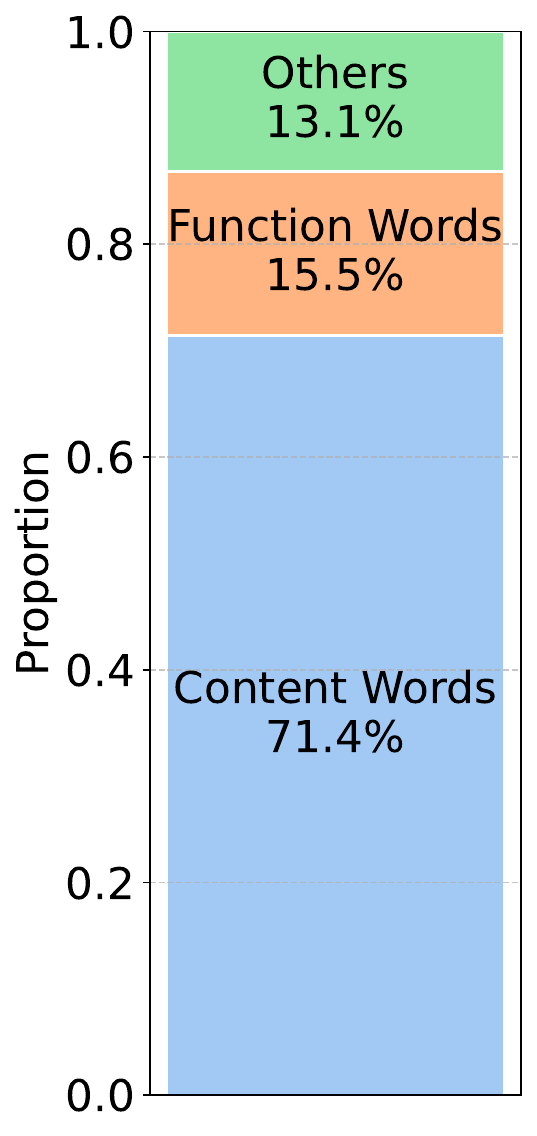}
        \caption{Qwen/Ko}
    \end{subfigure}

    \vspace{1em}

    \begin{subfigure}[b]{0.32\columnwidth}
        \centering
        \includegraphics[width=\linewidth]{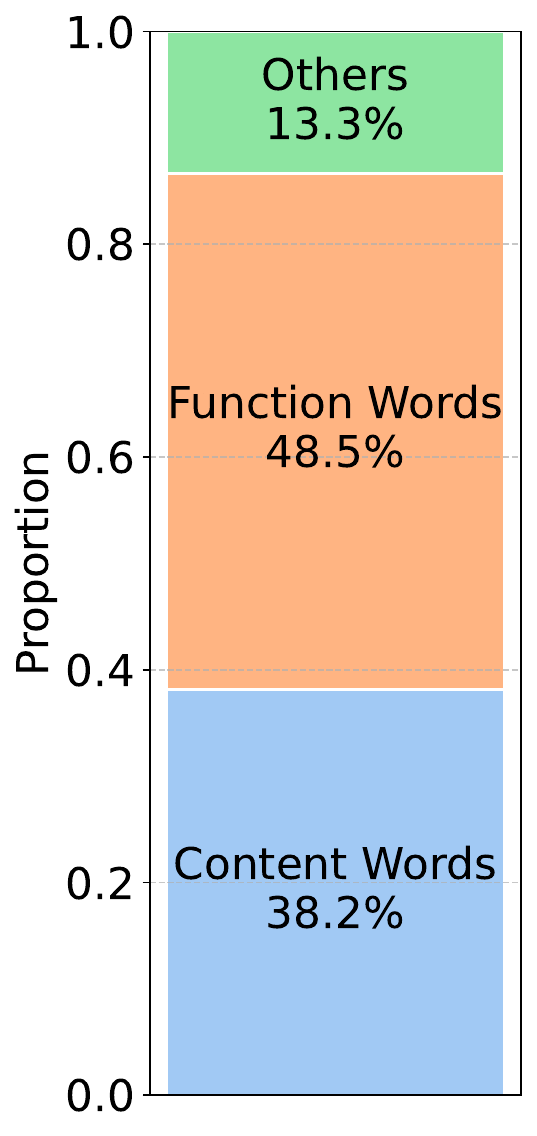} 
        \caption{CLOVA/En}
    \end{subfigure}
    \hfill 
    \begin{subfigure}[b]{0.32\columnwidth}
        \centering
        \includegraphics[width=\linewidth]{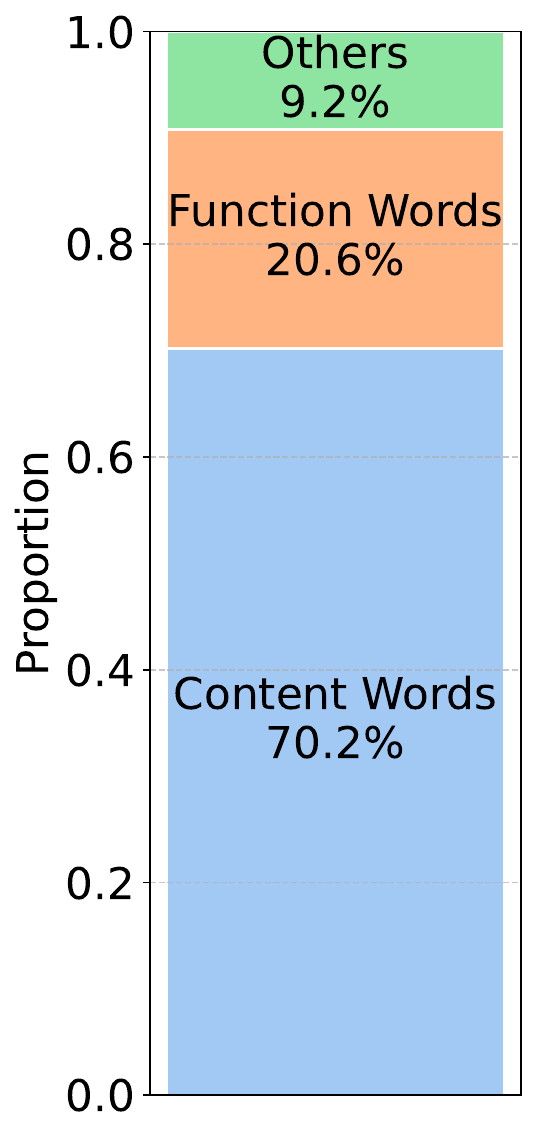} 
        \caption{CLOVA/Zh}
    \end{subfigure}
    \hfill 
    \begin{subfigure}[b]{0.32\columnwidth}
        \centering
        \includegraphics[width=\linewidth]{figures/z_score/k4_ud_korean_hyperclovax-0.5B.pdf}
        \caption{CLOVA/Ko}
    \end{subfigure}

    \caption{Proportional contribution of word categories to the total watermark detection z-score from STELA. 
    % Each subplot corresponds to a specific LLM~(rows) and language~(columns).
    }
    \label{fig:z_score_analysis_appendix}
\end{figure}

\section{Decomposing the Watermark Signal}
\label{appendix:decomposing_z_score}

Figure~\ref{fig:z_score_analysis_appendix} provides a comprehensive breakdown 
of the proportional contribution of different word categories to the total watermark 
detection z-score from STELA. 
The analysis spans all combinations of the models and languages tested 
in our experiments.
The central finding discussed in the main paper is 
consistent across all models. 
For English, the z-score contribution is nearly balanced between content words and 
function words, regardless of the model used for generation. 
In contrast, for both Chinese and Korean, the contribution from content words is 
consistently dominant across all three models. 
This demonstrates the generalizability of our linguistic analysis, 
showing that the watermark distribution is primarily determined by the 
typological characteristics of the language rather than the specific architecture 
of the language model.

\section{Typological Diversity}
\label{appendix:linguistic_background}

The application of syntactic predictability in STELA 
for watermarking requires empirical validation across typologically diverse languages.
Since the encoding of grammar differs fundamentally across language families, 
this variation creates a distinctive challenge for any single natural language processing system. 
We validate the approach on three languages that represent divergent structural frameworks: 
analytic English, isolating Chinese, and agglutinative Korean. 
An examination of their differences demonstrates the level of robustness 
required for a watermarking method to achieve broad generalizability.

\paragraph{Analytic English.}
English is a representative analytic language, where grammatical relationships 
are primarily determined by word order and the use of function words~\citep{comrie1989}. 
It adheres to a relatively strict Subject-Verb-Object~(SVO) word order to define the roles of 
sentence constituents. 
Function words, such as prepositions~(\textit{in, on, with}) and 
auxiliary verbs~(\textit{will, have, be}), are critical for conveying precise grammatical meanings. 
Compared to other language types, English has limited inflectional morphology; 
for instance, verbs and nouns undergo minimal form changes to indicate tense or number. 
Consequently, its part-of-speech system features distinct tags for these function words, 
such as prepositions~(IN) and determiners~(DT).

\paragraph{Isolating Chinese.}
Chinese exemplifies an isolating language, a purer form of the analytic type, 
where words are generally single, invariable morphemes~\citep{li-thompson-1981}. 
Grammatical meaning is conveyed almost exclusively through strict word order 
and the use of functional particles. 
Chinese has a near-total absence of inflectional morphology; 
there are no verb conjugations for tense or person, nor noun declensions for case. 
In compensation for the absence of inflectional morphology, 
the Chinese POS system incorporates distinctive functional categories, 
most prominently classifiers or measure words~(\textit{e.g., 个, 本}) 
indispensable for quantification, along with aspectual and 
structural particles~(\textit{e.g., 的, 了, 着}).

\paragraph{Agglutinative Korean.}
In contrast, Korean is a canonical agglutinative language, 
where grammatical relationships are primarily expressed by attaching various affixes 
to a root word~\citep{sohn2001}. 
Each affix, or morpheme, typically has a single, clear grammatical function. 
This is most prominent in its use of case markers known as particles~(\textit{josa}), 
which attach to nouns to explicitly mark their 
grammatical roles~(e.g., subject marker \textit{-이/-가}, object marker \textit{-을/-를}). 
Similarly, verb stems combine with a rich set of endings~(\textit{eomi}) to indicate tense, mood, 
and politeness levels. 
The richness of the morphological system gives rise to a notable flexibility in word order, 
in which Subject–Object–Verb~(SOV) functions as the canonical configuration yet 
does not operate as an obligatory pattern.
The POS tagging system for Korean is therefore necessarily highly granular, 
containing dozens of specific tags for these crucial functional morphemes.

% New Camera-ready
\section{Scalability to Larger Models}
\label{appendix:scalability}

We evaluate detection performance on Qwen3-4B, 
approximately $7{\times}$ larger than Qwen3-0.6B, 
in order to verify that STELA generalizes beyond the sub-1B 
parameter regime. 
We report results under both POS tagsets 
used in the main experiments: 
language-specific~(Table~\ref{tab:scalability_lang}) 
and universal dependencies~(Table~\ref{tab:scalability_ud}).

\begin{table}[hbt!]
\centering
\small
\begin{tabular}{ll|cc|c}
\toprule
\textbf{Lang.} & \textbf{Model} 
  & \textbf{TPR@5\%} & \textbf{Best F1} 
  & \textbf{$\Delta$TPR} \\
\midrule
\multirow{2}{*}{En} 
  & Qwen3-0.6B & 0.978 & 0.966 & -- \\
  & Qwen3-4B   & \textbf{0.996} & \textbf{0.980} & +1.8\% \\
\midrule
\multirow{2}{*}{Zh} 
  & Qwen3-0.6B & 0.996 & 0.994 & -- \\
  & Qwen3-4B   & \textbf{0.998} & \textbf{0.995} & +0.2\% \\
\midrule
\multirow{2}{*}{Ko} 
  & Qwen3-0.6B & 0.950 & 0.950 & -- \\
  & Qwen3-4B   & \textbf{0.980} & \textbf{0.967} & +3.0\% \\
\bottomrule
\end{tabular}
\caption{Scalability from Qwen3-0.6B to Qwen3-4B~(language-specific POS tagset).}
\label{tab:scalability_lang}
\end{table}

\begin{table}[hbt!]
\centering
\small
\begin{tabular}{ll|cc|c}
\toprule
\textbf{Lang.} & \textbf{Model} 
  & \textbf{TPR@5\%} & \textbf{Best F1} 
  & \textbf{$\Delta$TPR} \\
\midrule
\multirow{2}{*}{En} 
  & Qwen3-0.6B & 0.972 & 0.961 & -- \\
  & Qwen3-4B   & \textbf{0.994} & \textbf{0.976} & +2.2\% \\
\midrule
\multirow{2}{*}{Zh} 
  & Qwen3-0.6B & 0.998 & 0.993 & -- \\
  & Qwen3-4B   & \textbf{0.998} & 0.983         & 0.0\% \\
\midrule
\multirow{2}{*}{Ko} 
  & Qwen3-0.6B & 0.932 & 0.946 & -- \\
  & Qwen3-4B   & \textbf{0.968} & \textbf{0.961} & +3.6\% \\
\bottomrule
\end{tabular}
\caption{Scalability from Qwen3-0.6B to Qwen3-4B~(universal POS tagset).
}
\label{tab:scalability_ud}
\end{table}

As shown in Tables~\ref{tab:scalability_lang} 
and~\ref{tab:scalability_ud}, detection performance improves or 
remains near-saturated as model size 
increases, 
consistently across all three languages and both tagsets. 
Gains are most pronounced for 
Korean~(+3.0\% to +3.6\% TPR) and 
English~(+1.8\% to +2.2\% TPR), whereas 
Chinese performance is already near saturation at the 0.6B 
scale. 
We attribute these gains to the tendency of larger 
models to produce text that more closely adheres to natural 
syntactic patterns, which in turn renders the linguistic 
indeterminacy signal more salient and reliable. 

% New Camera Ready
\section{Correlation with Token Entropy}
\label{appendix:correlation}

A natural question is whether the linguistic indeterminacy 
signal $\lambda(c_t)$ meaningfully reflects model-internal 
uncertainty. 
We empirically address this question by computing the sentence-level 
Pearson correlation between $\lambda$ and token entropy on 
generated text.

We generate 500 English texts using Llama-3.2 with 
STELA watermarking. 
For each text, we compute 
1) the sentence-wise average linguistic indeterminacy, and
2) the sentence-wise average token entropy.
As shown in Table~\ref{tab:correlation}, we observe a 
statistically 
significant positive correlation~($r = 0.351$, $p < 10^{-15}$), indicating that $\lambda$ 
captures a meaningful component of the generative uncertainty 
of the model.

\begin{table}[hbt!]
\centering
\small
\begin{tabular}{cc}
\toprule
\textbf{Pearson $r$} & \textbf{$p$-value} \\
\midrule
0.351 & $6.33e-16$ \\
\bottomrule
\end{tabular}
\caption{Sentence-level Pearson correlation between average 
linguistic indeterminacy and average token 
entropy on 500 English texts generated by Llama-3.2.}
\label{tab:correlation}
\end{table}

Importantly, unlike token entropy, which is private, 
model-specific, and sensitive to 
decoding hyperparameters~(temperature, top-$k$/top-$p$, etc.), $\lambda$ is a stable, 
model-agnostic signal derived from the intrinsic syntactic 
structure of the language. 
This model-invariant property 
enables STELA to serve as a publicly verifiable proxy for 
adaptivity, a capability that entropy-based detectors 
inherently lack. 
The moderate magnitude of the correlation~($r \approx 0.35$ rather than near 1) is, in fact, a 
desirable property rather than a limitation: $\lambda$ 
captures syntactic flexibility specifically, whereas 
token entropy conflates syntactic, semantic, and stylistic 
uncertainties into a single undifferentiated signal. 

% New Camera-ready
\section{Empirical Null Distribution Analysis}
\label{appendix:empirical_null}

We conduct an empirical null analysis to assess whether the weighted 
$z$-test of STELA introduces systematic bias or variance 
mis-specification relative to the unweighted KGW baseline. 
Specifically, we generate 
1,500 non-watermarked texts~(500 per language) using 
Llama-3.2~(English), Qwen-3~(Chinese), 
and HyperCLOVA~(Korean), and compute the $z$-score distributions 
under both STELA and KGW.

\begin{table}[hbt!]
\centering
\small
\begin{tabular}{ll|cc}
\toprule
\textbf{Language} & \textbf{Method} 
  & \textbf{Mean ($\mu$)} & \textbf{Std ($\sigma$)} \\
\midrule
\multirow{2}{*}{English}
  & KGW   & 0.25 & 1.12 \\
  & STELA & 0.35 & 1.17 \\
\midrule
\multirow{2}{*}{Chinese}
  & KGW   & 0.24 & 2.22 \\
  & STELA & 0.25 & 2.19 \\
\midrule
\multirow{2}{*}{Korean}
  & KGW   & \phantom{$-$}0.03 & 2.95 \\
  & STELA & $-$0.10 & 3.02 \\
\bottomrule
\end{tabular}
\caption{Empirical null $z$-score distributions computed over 
1,500 non-watermarked texts. The adaptive weighting of STELA 
does not exacerbate variance relative to the unweighted KGW baseline.}
\label{tab:empirical_null}
\end{table}

As shown in Table~\ref{tab:empirical_null}, 
for English, both methods closely approximate $\mathcal{N}(0,1)$, 
confirming the validity of the token independence assumption. 
For Chinese and Korean, we observe elevated variance under both methods, 
which reflects stronger inter-token dependencies inherent to these 
languages rather than any deficiency in our weighting scheme. 
Critically, STELA does not amplify this deviation: in 
Chinese, STELA attains $\sigma \approx 2.19$, which is marginally 
lower than that of KGW ($\sigma \approx 2.22$). 
This confirms the statistical validity of our weighted z-test.

\end{document}